\documentclass[10pt,twocolumn,letterpaper]{article}

\usepackage{cvpr}              

\usepackage{subcaption}
\usepackage{amsmath,amsfonts,bm}
\usepackage{amssymb}
\usepackage{booktabs,siunitx,multirow}
\usepackage{colortbl}
\usepackage{float}
\usepackage{caption}
\usepackage{dblfloatfix} 
\usepackage{cuted}      
\usepackage{caption}    
\usepackage{graphicx}
\usepackage{draftfigure}
\usepackage{todonotes}
\usepackage{ifthen}

\definecolor{cvprblue}{rgb}{0.21,0.49,0.74}
\usepackage[pagebackref,breaklinks,colorlinks,allcolors=cvprblue]{hyperref}
\usepackage{soul}
\usepackage{multirow}

\definecolor{2nd}{HTML}{FFECC0}
\definecolor{1st}{HTML}{FFC7A7}

\def\eqref#1{equation~\ref{#1}}

\def\1{\bm{1}}

\def\rmT{{\mathbf{T}}}

\def\vc{{\bm{c}}}

\def\vn{{\bm{n}}}
\def\vo{{\bm{o}}}
\def\vp{{\bm{p}}}
\def\vq{{\bm{q}}}

\def\vs{{\bm{s}}}

\def\mI{{\bm{I}}}

\def\mT{{\bm{T}}}

\DeclareMathAlphabet{\mathsfit}{\encodingdefault}{\sfdefault}{m}{sl}
\SetMathAlphabet{\mathsfit}{bold}{\encodingdefault}{\sfdefault}{bx}{n}

\def\gA{{\mathcal{A}}}

\def\gF{{\mathcal{F}}}

\def\gH{{\mathcal{H}}}

\def\gL{{\mathcal{L}}}

\newcommand{\R}{\mathbb{R}}

\def\paperID{34646} 
\def\confName{CVPR}
\def\confYear{2026}

\title{Large-scale Codec Avatars: \\The Unreasonable Effectiveness of Large-scale Avatar Pretraining}

\author{\parbox{\linewidth}{\centering\normalsize
Junxuan Li$^{*}$,
Rawal Khirodkar$^{*}$,
Chengan He,
Zhongshi Jiang,
Giljoo Nam,
Lingchen Yang,
Jihyun Lee,
Egor Zakharov,
Zhaoen Su,
Rinat Abdrashitov,
Yuan Dong,
Julieta Martinez,
Kai Li,
Qingyang Tan,
Takaaki Shiratori,
Matthew Hu,
Peihong Guo,
Xuhua Huang,
Ariyan Zarei,
Marco Pesavento,
Yichen Xu,
He Wen,
Teng Deng,
Wyatt Borsos,
Anjali Thakrar,
Jean-Charles Bazin,
Carsten Stoll,
Gin\'{e}s Hidalgo,
James Booth,
Lucy Wang,
Xiaowen Ma,
Yu Rong,
Sairanjith Thalanki,
Chen Cao,
Christian H\"{a}ne,
Abhishek Kar,
Sofien Bouaziz,
Jason Saragih,
Yaser Sheikh,
Shunsuke Saito$^{\dagger}$\\[4pt]
Codec Avatars Lab, Meta\\[2pt]
{\small\url{https://junxuan-li.github.io/lca}}
}}

\begin{document}
\maketitle

\renewcommand{\thefootnote}{\fnsymbol{footnote}}
\footnotetext[0]{$^{*}$Core contributors\hspace{1em}$^{\dagger}$Project lead}

\begin{strip}
    \centering
    \vspace{-0.8in}
    \begin{minipage}{\textwidth}
        \centering
        \includegraphics[width=\linewidth]{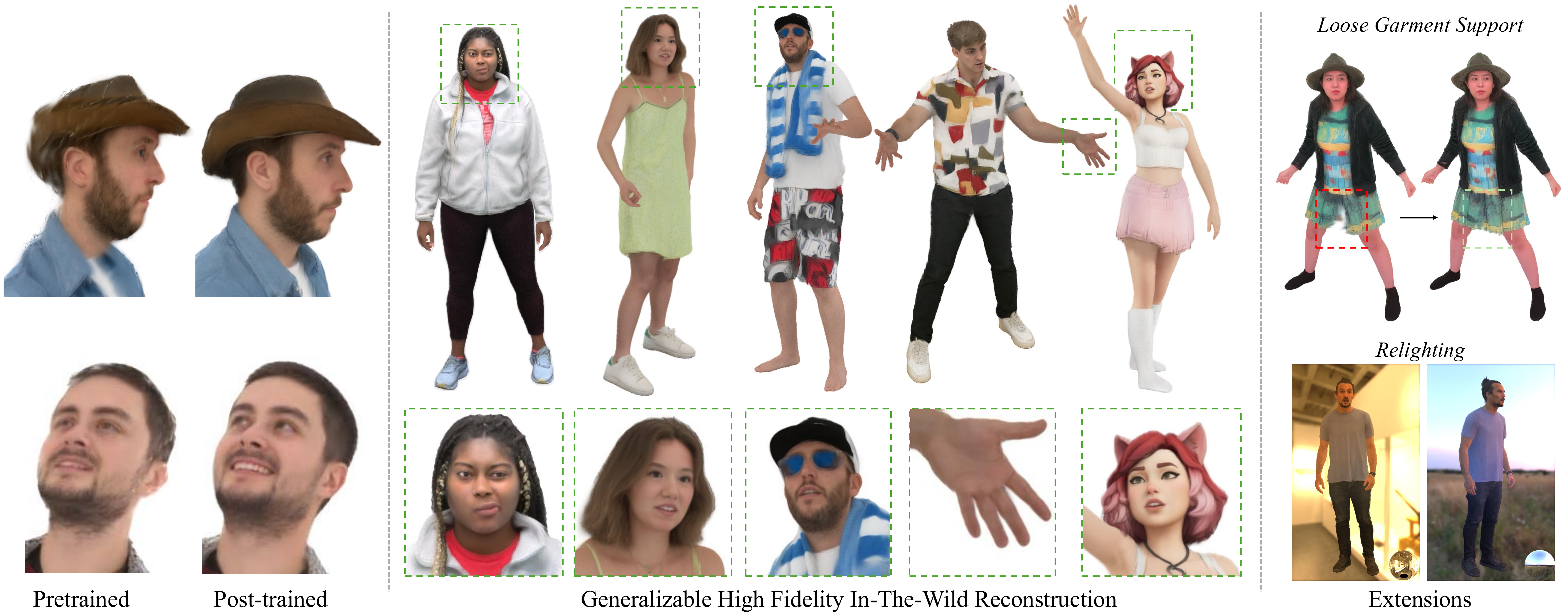}
        \captionof{figure}{\textbf{Large-scale Codec Avatars (LCA).} (\textit{Left}) We generate avatars from a handful of images in seconds. LCA follows the pre/post-training paradigm, achieving broad generalization with high-fidelity reconstruction over pretraining alone. (\textit{Middle}) The resulting avatars are highly detailed with faithful 3D structures, fully animatable with expression, gaze, and body pose, even for out-of-domain samples. (\textit{Right}) LCA further supports \textit{loose garments} and \textit{relighting} while retaining generalizability by only modifying the post-training.}
        \label{figure:teaser}
    \end{minipage}
\end{strip}

\begin{abstract}
High-quality 3D avatar modeling faces a critical trade-off between fidelity and generalization. On the one hand, multi-view studio data enables high-fidelity modeling of humans with precise control over expressions and poses, but it struggles to generalize to real-world data due to limited scale and the domain gap between the studio environment and the real world. On the other hand, recent large-scale avatar models trained on millions of in-the-wild samples show promise for generalization across a wide range of identities, yet the resulting avatars are often of low-quality due to inherent 3D ambiguities. To address this, we present Large-Scale Codec Avatars (LCA), a high-fidelity, full-body 3D avatar model that generalizes to world-scale populations in a feedforward manner, enabling efficient inference. Inspired by the success of large language models and vision foundation models, we present, for the first time, a pre/post-training paradigm for 3D avatar modeling at scale: we pretrain on 1M in-the-wild videos to learn broad priors over appearance and geometry, then post-train on high-quality curated data to enhance expressivity and fidelity. LCA generalizes across hair styles, clothing, and demographics while providing precise, fine-grained facial expressions and finger-level articulation control, with strong identity preservation. Notably, we observe emergent generalization to relightability and loose garment support to unconstrained inputs, and zero-shot robustness to stylized imagery, despite the absence of direct supervision.
\end{abstract}

\vspace{-0.2in}
\section{Introduction}
\label{section:introduction}

Photorealistic human avatars~\cite{wang2025relightable,saito2024relightable,qian2024gaussianavatars,zheng2023avatarrex,li2022tava,li2024uravatar,zielonka2025drivable} present the opportunity to transform how humans communicate~\cite{yu2021avatars}, but broad adoption requires systems that work robustly for everyone. We present an approach that, given a handful of images of a subject, produces an identity-preserving 3D avatar that can be accurately driven by subtle facial expressions, full-body motion, and fine-grained hand poses. Achieving this goal requires preserving two key properties that trade-off against each other:
(1) \textit{generalization} across clothing, hairstyles, accessories, demographics, and environments, and (2) \textit{fidelity} preserving precise motion and 3D consistent authenticity. 
Together, these properties are necessary for enabling a true communication service for everyone, as identity must be preserved no matter whom you interact with and the signal embedded in behavior and appearance must be preserved no matter what they do.

Most existing approaches trade generalization and fidelity against each other. One line of work uses \textbf{high-quality studio data}, with the number of identities typically in the thousands at most~\cite{cao2022authentic,saito2024relightable,xu2024gaussian,li2024uravatar,liu2025lucas,guo2025vid2avatar}, as they use expensive optimization pipelines for personalization. These systems deliver authentic and expressive avatars but they generalize poorly beyond the captured domains. Another line trains on diverse corpora~\cite{LHM,qiu2025pf,he2025lam} from \textbf{in-the-wild data}. These models generalize more broadly in a feedforward manner but often produce distortions from unobserved views, blur in body parts, and limited expressivity. 

Recently, large-scale pre/post-training has achieved remarkable success in resolving the aforementioned trade-off in language modeling~\cite{achiam2023gpt, touvron2023llama2,team2023gemini}, vision models~\cite{simeoni2025dinov3,bolya2025perception,khirodkar2024sapiens} and video generation~\cite{wan2025wan,bar2024lumiere,kong2024hunyuanvideo}. Pretraining learns broad priors for generalization from million-to-billion-scale training data, and the model is post-trained with high-quality curated data to align the learned representation with a target task. Inspired by the success in adjacent domains, we present Large-scale Codec Avatars (LCA), a pre/post-train framework for human avatar creation. LCA first \textit{pre-trains} on millions of in-the-wild videos to learn human priors over appearance and geometry for generalization. 
It then \textit{post-trains} on high-resolution, multi-view studio captures~\cite{martinez2024codec} spanning thousands of identities to specialize for precise control and photorealism. We show, for the first time, that this two-stage approach at scale breaks the generalization-fidelity trade-off, yielding avatars that are fully expressive, identity-preserving, and robustly generated under real-world conditions. \Cref{figure:teaser} qualitatively compares the two stages: the pretrained model generalizes across ethnicity, clothing, and hairstyles, but exhibits muted expressions with distorted 3D shapes, whereas the post-training produces more expressive facial animation with faithful 3D structure while preserving the identity.

Extending the pre/post-training paradigm to human avatar modeling poses a unique challenge: the architecture needs to be scalable, expressive, and efficient. To unify training with studio and in-the-wild data, we adopt a scalable architecture that implicitly associates one or more reference images of a subject with animatable 3D Gaussians~\cite{LHM}. Unlike prior methods based on studio data~\cite{li2024uravatar,guo2025vid2avatar}, our approach does not require high-quality conditioning data, such as geometry and texture maps, allowing seamless support of pre/post-training. To capture full-body expressivity, LCA uses a two-branch design: one outputs canonical appearance and geometry, and the other decodes correctives to the canonical output driven by body/hand poses of an expressive body model~\cite{park2025atlas} and facial expression latent codes learned in a self-supervised manner similar to~\cite{xu2024vasa}.

Specifically, we form two token streams: image tokens from off-the-shelf visual extractors~\cite{khirodkar2024sapiens} and geometric tokens from a template body mesh in a canonical pose. A large transformer~\cite{shin2025exploring} backbone fuses these tokens. To support a variable number of input images, we adopt a hybrid attention scheme that alternates global attention over all tokens with per-image self-attention blocks~\cite{wang2025vggt}.  A canonical MLP branch decodes the output tokens into per-Gaussian canonical attributes (center, rotation, scale, opacity, and color). A corrective MLP branch predicts per-Gaussian attribute offsets conditioned on the output tokens and the driving signals. 
The attributes with correctives are transformed to the target pose via linear blend skinning (LBS), and then rendered with differentiable 3D Gaussian splatting~\cite{kerbl20233d}.

The most remarkable characteristic of LCA is its strong generalizability. LCA faithfully reconstructs clothing, hairstyles, and accessories that do not exist in the post-training data (\eg, eyewear, headwear).  
We also show that LCA can easily incorporate additional features such as loose-garment handling and relighting~\cite{wang2025relightable} while retaining its generalizability to unconstrained inputs by only modifying the post-training stage. Moreover, LCA generalizes to stylized or fictional characters despite explicitly filtering them out from both pre/post-training. 
Our experiments show that LCA sets a new state-of-the-art in avatar modeling and faithfully captures subtle facial expressions and whole-body motion, including finger-level articulation. Finally, its modular design enables avatar creation in seconds and real-time animation: the pose-dependent residual head is lightweight and runs per frame, while the transformer inference is executed only once during generation.

\noindent
In summary, our contributions are as follows:
\begin{itemize}
\itemsep0em
  \item We are the first to show that million-scale pre/post-training simultaneously yields broad generalization with high-fidelity outputs for animatable avatar creation.
  \item We propose a new architecture that supports flexible identity conditioning while supporting faithful facial and whole-body animation.
  \item Our core design is versatile and efficient: LCA extends to additional features, including loose garment handling and relighting, with minimal modifications, and the avatars can be animated in real-time.
\end{itemize}

\section{Related Work}
\label{section:related_work}

\textbf{Studio-Based 3D Avatars.} 
3D human avatar modeling has been actively studied over the past decades~\cite{wang2024survey}, and the data available for avatar creation has been a key factor affecting the fidelity of the resulting avatars. The line of work achieving the highest quality typically relies on multi-view studio data captured in calibrated, highly controlled environments~\cite{sun2020light,alexander2010digital,martinez2024codec,qian2024gaussianavatars,bagautdinov2021driving,chen2024meshavatar,zheng2023avatarrex,li2022tava,li2024uravatar,li2024animatable,zielonka2025drivable,ma2021pixel,saito2024relightable}. Such setups provide dense observations of the identity across diverse viewpoints, appearances, and motions, enabling the effective learning of 3D avatar representations (e.g., NeRF~\cite{mildenhall2021nerf,peng2021animatable,NerFACE}, 3DGS~\cite{qian2024gaussianavatars,saito2024relightable,qian2024gaussianavatars}) to achieve high authenticity and expressiveness. Relightability is another core capability that can be effectively learned in studio settings equipped with light stages~\cite{guo2019relightables,bi2021deep,yang2023towards,saito2024relightable,he2024diffrelight,li2024uravatar,wang2025relightable}, which are essential for achieving photorealistic appearance under varying illumination. Despite their remarkable quality, acquiring calibrated multi-view captures is impractical for  users, and these methods often perform poorly when directly generalized to in-the-wild inputs due to domain gap.

\vspace{1mm}\noindent
\textbf{In-the-Wild 3D Avatars.}
Unlike studio-based avatars, approaches that create avatars from in-the-wild (ITW) data reflect more practical real-world scenarios. Most existing methods either (1) learn a feedforward 3D avatar reconstruction model from large-scale, casually captured images or videos~\cite{LHM,zhuang2025idol,weng2024template}, or (2) optimize 3D avatar representations directly from ITW captures~\cite{moon2024expressive,jiang2023instantavatar,guo2023vid2avatar,sim2025persona,jiang2022neuman}.
However, the avatar creation problem in this setting remains highly under-constrained for achieving high \textit{3D} fidelity, as ITW captures typically provide only sparse and monocular observations.
To mitigate this, recent methods attempt to reduce the 3D ambiguity by (1) leveraging image generative models to augment ITW observations~\cite{weng2024template,sim2025persona}, or (2) learning a universal prior model additionally trained on multi-view data~\cite{li2024uravatar,guo2025vid2avatar}. Nevertheless, these approaches still require expensive test-time fine-tuning on ITW captures to achieve reasonable quality -- falling short of the fidelity and authenticity attained by avatars created from studio-captured data. In summary, studio-based avatars achieve high fidelity but lack generalizability, whereas in-the-wild avatars exhibit the opposite trade-off. To bridge this gap, we incorporate pre/post-training for 3D avatar modeling that jointly leverages the advantages of both data regimes.

\vspace{1mm}\noindent
\textbf{Large-Scale Pre/Post-Training.}
Beyond the 3D avatar domain, recent large language models (LLMs)~\cite{touvron2023llama,devlin2018bert,tay2021scale,achiam2023gpt,touvron2023llama2,team2023gemini,jiang2023mistral,brown2020language} and image or video generative models~\cite{wan2025wan,bar2024lumiere,kong2024hunyuanvideo,bruce2024genie} have demonstrated remarkable performance, achieving both high fidelity and strong generalization. This success largely stems from a two-stage learning paradigm comprising \emph{pretraining} and \emph{post-training}. In the \emph{pretraining stage}, models are trained on massive, diverse datasets to learn comprehensive inductive priors without focusing on specific downstream objectives. While this stage provides robust generalization, it often yields suboptimal fidelity due to noisy, heterogeneous data. In the subsequent \emph{post-training stage}, the model is fine-tuned on smaller, high-quality data to enhance fidelity, alignment, and controllability. For example, recent LLMs~\cite{touvron2023llama,devlin2018bert,tay2021scale,achiam2023gpt,touvron2023llama2,team2023gemini,jiang2023mistral,brown2020language} are pre-trained on trillions of internet-scale text tokens and then post-trained to align with human preferences (e.g., RLHF~\cite{ouyang2022training}, DPO~\cite{rafailov2023direct}). Similarly, modern image and video generative models~\cite{wan2025wan,bar2024lumiere,kong2024hunyuanvideo,bruce2024genie} are first pre-trained on large-scale visual data and later post-trained for higher fidelity or controllability~\cite{cheng2025wan,hu2024animate,kim2025personabooth}.
Despite its demonstrated effectiveness in other domains, large-scale pre- and post-training have not yet been explored for 3D avatar modeling---a direction we argue is crucial for achieving fidelity and generalization.

\begin{figure*}[t]
\centering
\includegraphics[width=\linewidth]{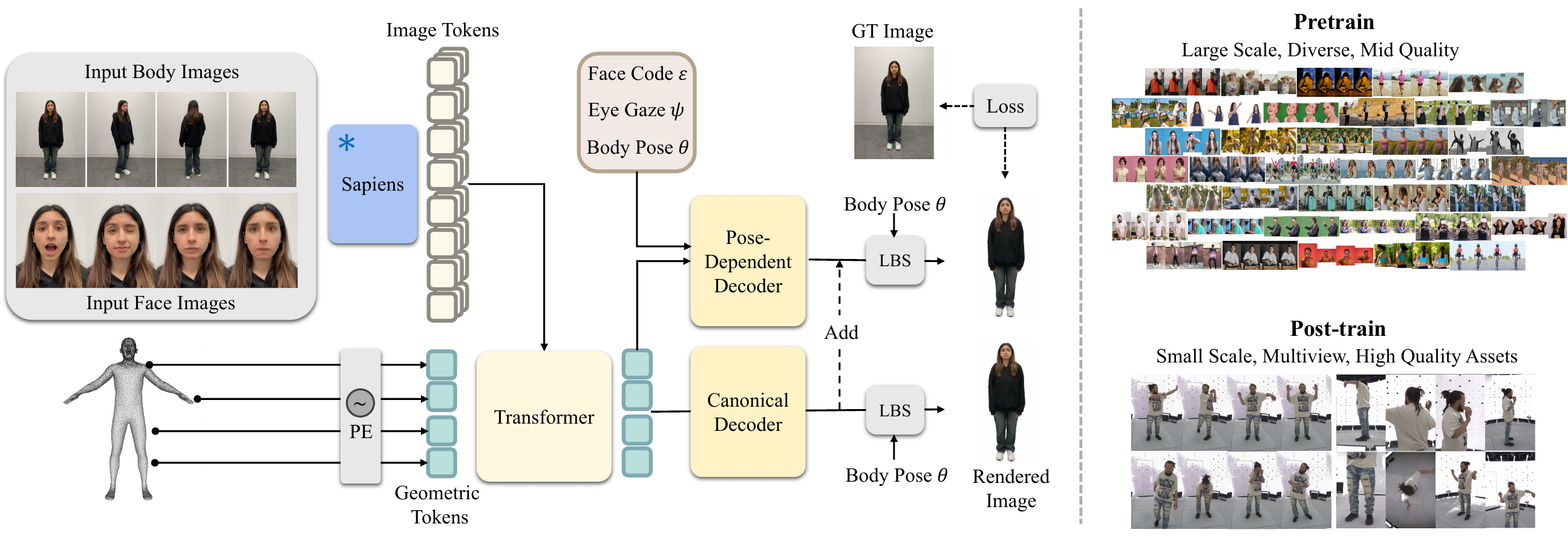}
\caption{(\textit{Left}) \textbf{Overview.} Given multiple images of a subject, we extract \emph{image tokens} from full-body images and face crops, and \emph{geometric tokens} from a template mesh. The LCA encoder alternates image-only, geometry-only, and multimodal attention to fuse information across streams. Our decoders, canonical and pose-dependent, predict Gaussian attributes, which are skinned via linear blend skinning (LBS) and rendered to novel views. Training uses photometric reconstruction losses. (\textit{Right}) \textbf{Pretraining vs.\ Post-Training.} LCA pretrains on large-scale, unconstrained monocular videos of single subjects with mixed (mid/low) quality, then post-trains on high-quality, multi-view studio captures. Pretraining drives broad generalization whereas post-training improves fidelity and 3D completeness. }

\vspace*{-0.2in}
\label{figure:method}
\end{figure*}

\section{Large-scale Codec Avatars}
\label{section:method}

In this section, we detail the architecture (\Cref{sec:architecture}), objective (\Cref{sec:loss}), data preparation and pre/post-training setup(\Cref{sec:data}), and feature extensions (\Cref{sec:extension}).

\subsection{Architecture}
\label{sec:architecture}
\noindent\textbf{Tokenization.} 
Given $N$ images of the full-body $\mathbf{I}_i^{\text{body}}= \{\mathbf{I}_i \in \mathbb{R}^{H\times W \times 3}\}_{i=1}^{N}$ and face close-up $\mathbf{I}_i^{\text{face}}$,
we compute their image features with Sapiens~\cite{khirodkar2024sapiens}, denoted $\mathcal{E}_{\mathrm{sap}}$, and mapped to a $D$-dimensional token space by a shared single-layer MLP $\mathcal{F}_{\mathrm{proj}}$:

\vspace{-0.2in}
\begin{align}
\mathbf{T}_i^{\text{body}}  = \mathcal{F}_{\text{proj}}(\mathcal{E}_{\text{sap}} (\mathbf{I}_i^{\text{body}})),\\
\mathbf{T}_i^{\text{face}}  = \mathcal{F}_{\text{proj}}(\mathcal{E}_{\text{sap}} (\mathbf{I}_i^{\text{face}})),
\end{align}
where $\mathbf{T}_i^{\text{body}}, \mathbf{T}_i^{\text{face}} \in \mathbb{R}^{P\times D}$ with $P$ denoting the number of patches. In addition, we sample $G$ anchor points from a template 3D human mesh with positions $\mathbf{X} \in \R^{G \times 3}$, following LHM~\cite{LHM}. These are encoded by a positional encoder $\gF_{\text{PE}}$ and projected to the same hidden dimension via $\mathcal{F}_{\text{proj-gs}}$ to give $\mathbf{T}^{\text{gs}} \in \mathbb{R}^{G\times D}$:
\begin{align}
\mathbf{T}^{\text{gs}} =\mathcal{F}_{\text{proj-gs}} (\mathcal{F}_{\text{PE}} (\mathbf{X})).
\end{align}
Together, these form two token streams: (i) image tokens $\{\mathbf{T}_i^{\text{body}}, \mathbf{T}_i^{\text{face}}\}_{i=1}^N$ and (ii) geometric tokens $\mathbf{T}^{\text{gs}}$. 

\vspace{1mm}\noindent\textbf{Transformer.}
To efficiently share information across the two token streams, each LCA encoder layer consists of three stages: (i) image attention—self-attention among the image tokens (ii) geometric attention—self-attention among the geometric tokens and (iii) multimodal attention—self-attention over the concatenated image and geometric tokens to fuse body, face, and geometric cues. For each view 
$i$,
\begin{align}
\mathbf{T}^{\text{body}}_i =  \mathcal{A}_{\text{image}}(\mathbf{T}^{\text{body}}_{i}), \\
\mathbf{T}^{\text{face}}_i =  \mathcal{A}_{\text{image}}(\mathbf{T}^{\text{face}}_{i}),
\end{align}
where $\mathcal{A}_{\text{image}}$ applies self-attention to the image tokens independently, together with standard operations such as LayerNorm~\cite{xu2019understanding}, residual connections, and MLPs. Similarly, we apply geometric attention using $\mathcal{A}_{\text{geometry}}$ on 

\vspace{-0.1in}
\begin{align}
\mathbf{T}^{\text{gs}} = \mathcal{A}_{\text{geometry}}(\mathbf{T}^{\text{gs}}).
\end{align}

We then concatenate per-view outputs, $\mathbf{T}^{\text{body}} = \big[\mathbf{T}^{\text{body}}_1,\ldots,\mathbf{T}^{\text{body}}_N\big]$, $\mathbf{T}^{\text{face}} = \big[\mathbf{T}^{\text{face}}_1,\ldots,\mathbf{T}^{\text{face}}_N\big]$ and perform the multimodal attention using $\mathcal{A}_{\text{multimodal}}$,
\begin{align}
\mathbf{T}^{\text{gs}}, \mathbf{T}^{\text{body}}, \mathbf{T}^{\text{face}} = \mathcal{A}_{\text{multimodal}} (\mathbf{T}^{\text{gs}}, \mathbf{T}^{\text{body}}, \mathbf{T}^{\text{face}}).
\end{align}
$\mathcal{A}_{\text{multimodal}}$ architecturally resembles the body–face MMDiT~\cite{esser2024scaling} block of LHM~\cite{LHM}, which uses masked attention so that only face geometric tokens attend to face image tokens. Together $\mathcal{A}_{\text{image}}, \mathcal{A}_{\text{geometry}}, \mathcal{A}_{\text{multimodal}}$ operations constitute a single layer among the $L$ layers of our encoder. This design supports an arbitrary number of input views~\cite{wang2025vggt} and enables bidirectional information exchange between image and geometric tokens.

\vspace{1mm}\noindent\textbf{Gaussian Decoder.}
We decode the geometric tokens into 3D Gaussian attributes -- position, rotation, scale, opacity, and color. Our decoder has two heads: a canonical head to capture static features, and a pose-dependent head to model pose-driven effects such as facial expressions, eye gaze, hand pose and clothing deformations.

\vspace{1mm}\noindent
\textit{Canonical:} The encoded geometric tokens $\mathbf{T}^{\text{gs}}$ are decoded into canonical 3D Gaussians using an MLP head $\mathcal{H}_\text{cano}$,
\begin{align}
\vc, \vp, \vo, \vq, \vs = \mathcal{H}_{\text{cano}}(\mathbf{T}^{\text{gs}}),
\end{align}
where $\vc \in \R^{kG \times 3}, 
\vp \in \R^{kG \times 3}, \vo \in \R^{kG}, \vq \in \R^{kG \times 4}, \vs \in \R^{kG \times 3}$ denote the color, position, opacity, quaternion rotation, and scale of each 3D Gaussian, respectively. Note that $k$ is a Gaussian-to-token ratio: each geometric token is expanded into $k$ distinct Gaussians by $\mathcal{H}_{\text{cano}}$.

\vspace{1mm}\noindent
\textit{Pose-Dependent:} Given body pose~\cite{park2025atlas} $\boldsymbol{\theta} \in \R^{138}$, face expression code $\boldsymbol{\varepsilon} \in \R^{128}$, and gaze direction $\boldsymbol{\psi} \in \R^6$, we concatenate these driving signals with the geometric tokens $\mathbf{T}^{\text{gs}}$ and pass them to an MLP head $\mathcal{H}_{\text{pose}}$ to predict the pose- and expression-dependent deltas of the Gaussian attributes:
\begin{align}
\Delta \vc,\Delta \vp, \Delta \vq, \Delta \vs = \mathcal{H}_{\text{pose}} (\mT^{\text{gs}} , \boldsymbol{\theta}, \boldsymbol{\varepsilon}, \boldsymbol{\psi}).
\end{align}
We apply these deltas to the canonical attributes $(\vc, \vp, \vo, \vq, \vs)$ to obtain pose- and expression-aware Gaussians, while keeping opacities $\vo$ fixed during animation to promote stability across poses and expressions. \Cref{figure:method} illustrates our LCA architecture overview.

\subsection{Loss}
\label{sec:loss}
Our training objective combines a photometric rendering loss with Gaussian regularizations.  We transform the canonical and pose-dependent Gaussian attributes to the target view using linear blend skinning (LBS) and render the canonical image $\hat{\mI}_{\text{cano}}$ and the pose-dependent image $\hat{\mI}_{\text{pose}}$. We supervise both renderings with $\ell_1$ and LPIPS~\cite{zhang2018unreasonable} loss,
\begin{align}
  \gL_{\text{img}}(\mI,\hat{\mI})
= \gL_{\ell_1}(\mI,\hat{\mI}) + \gL_{\text{LPIPS}}(\mI,\hat{\mI}).
\end{align}
We regularize Gaussian positions $\vp$ and scales $\vs$ as,
\begin{align}
\gL_{\text{reg}}(\vp,\vs)
= \gL_{\text{ACAP}}(\vp) + \gL_{\text{ASAP}}(\vs),
\end{align}
where $\gL_{\text{ACAP}}$ and $\gL_{\text{ASAP}}$ are position- and scale-regularizers~\cite{LHM}. The total loss per training sample is,
\begin{align}
\gL = \gL_{\text{img}}\big(\mI,\hat{\mI}_{\text{cano}}\big)+ \gL_{\text{img}}\big(\mI,\hat{\mI}_{\text{pose}}\big) + \lambda\gL_{\text{reg}}(\vp,\vs),
\end{align}
where $\lambda$ is the regularization weight. We observe that adding the photometric rendering loss explicitly against $\hat{\mI}_{\text{cano}}$ leads to faster convergence.

\subsection{Pretraining and Post-Training}
\label{sec:data}
We use distinct data sources for LCA’s two-stage training. Fig.~\ref{figure:method} (\textit{Right}) contrasts the data sources used in each stage.

\vspace{1mm}\noindent
\textbf{Pretraining.} We curated an in-the-wild dataset of 1 million monocular, human-centric videos. Each video contains a single subject and has a minimum diagonal resolution of $256$ pixels. In addition, we collect $40{,}000$ upper-body videos with diverse facial expressions, and full-body videos from $\sim\!1{,}000$ subjects performing a broad range of motions, captured from diverse viewpoints. During pretraining, both the encoder and decoder are randomly initialized and trained with a higher learning rate. This stage builds broad generalization to diverse inputs.

\vspace{1mm}\noindent
\textbf{Post-Training.} We use the term \emph{post-training} to refer to supervised fine-tuning on a small, high-quality dataset to improve avatar animatability and visual fidelity. We use a multi-view capture system~\cite{martinez2024codec} to record dynamic human performances. The setup uses $200$ calibrated, synchronized cameras capturing 4K images. Participants perform casual motions, yielding on average $\sim\!5{,}000$ frames per subject. In total, we collect recordings from $2{,}737$ participants for model training. During post-training, we start from the pretrained checkpoint and apply layer-wise learning-rate decay to preserve knowledge acquired during pretraining. Training on this high-quality multi-view data refines the model, improving 3D completeness and fine-grained details.

\subsection{Post-Training Extensions}
\label{sec:extension}
LCA architecture is versatile and extends to multiple applications with minimal modifications in the post-training.

\vspace{1mm}\noindent
\textbf{Loose Garment Support.}
Methods that use predefined skinning weights often produce garment-splitting artifacts when animating loose garments (\eg skirts)~\cite{LHM,qiu2025pf}. Nevertheless, 
we adopt such a conventional approach during pre-training for scalability. Specifically, given a predefined skinning weight field $\mathcal{W}$, we deform each Gaussian as
\begin{align}
     \hat{\vp} = \text{LBS}( \boldsymbol{\theta}, \vp + \Delta\vp; \mathcal{W}).
\end{align}

\begin{figure}[t]
\centering
\includegraphics[width=0.9\linewidth]{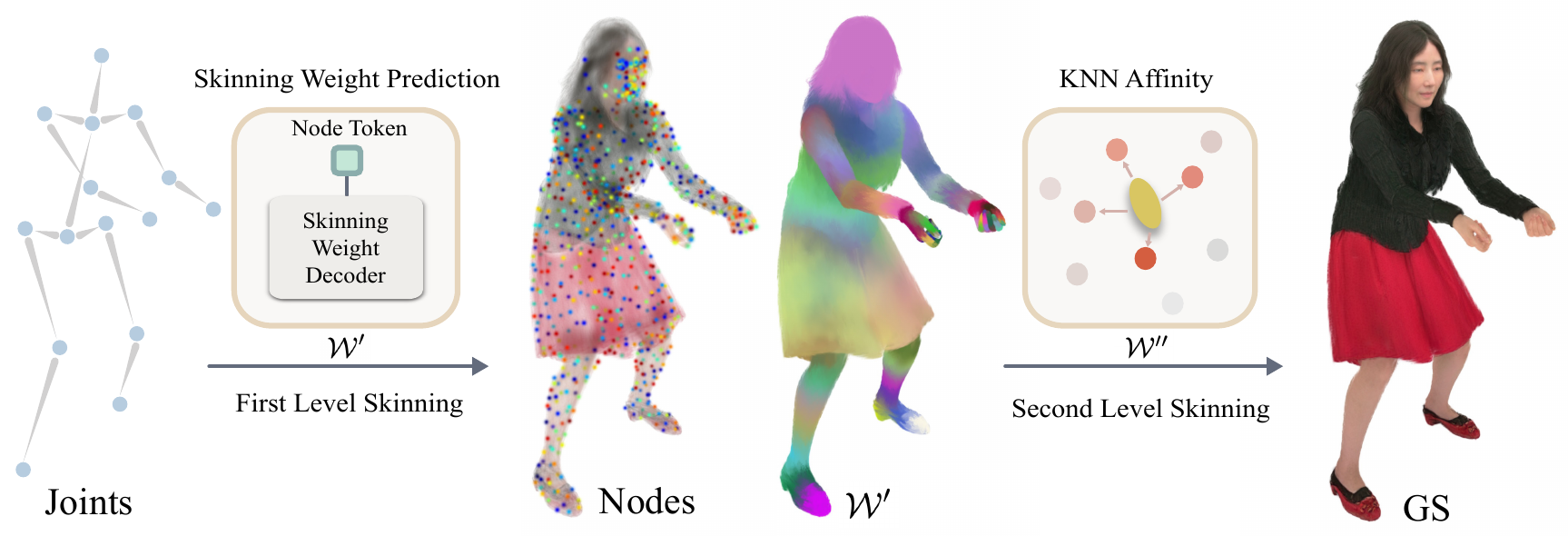}
\caption{\textbf{Node-Based Deformation Model.} 
We use a flexible two-level learnable deformation model to adapt skinning weights learning for post-training.}
\vspace*{-0.2in}
\label{figure:deformer}
\end{figure}

Since such fixed skinning weights cannot account for the large variation introduced by clothed humans, we enable loose-garment support in post-training stage. Inspired by \cite{sumner2007embedded,der2006inverse,pan2022predicting,guo2024reloo}, we introduce a two-level (coarse-to-fine) learnable deformation module. We define a set of intermediate nodes $\vn \in \R^{N_{\text{node}} \times 3}$ to encode a low-dimensional deformation subspace articulated by $\boldsymbol{\theta}$ through node-level skinning weights $\mathcal{W}'$. These nodes drive full-Gaussian deformation via embedded deformation weights $\mathcal{W}''$~\cite{sumner2007embedded} using 4-nearest-neighbors (\Cref{figure:deformer}). To enable subject-specific variation, we parameterize this subspace using learnable correctives applied on the spatial canonical weights~\cite{lin2022learning}:
\begin{align}
     \mathcal{W}' = \mathcal{W}(\vn) + \mathcal{H}_\text{skin}(\mathbf{T}^{\text{node}}),
\end{align}
where $\mathcal{H}_\text{skin}$ is an MLP head. For simplicity, 
node locations $\vn$ and tokens $\mathbf{T}^{\text{node}} \in \R^{N_{\text{node}} \times D}$  are uniformly sub-sampled from the canonical Gaussians.
To learn $\mathcal{H}_\text{skin}$, we add regularizations to encourages smooth yet sparse correctives:
\begin{align}
\gL_{\text{skin}}
= \gL_{\text{ARAP}}( \hat{\vp}, \vp)
+ \lambda_{\text{skw}}\gL_{\ell_1}(\mathcal{H}_{\text{skin}}),
\end{align}
where $\gL_{\text{ARAP}}$ is the As-Rigid-As-Possible loss~\cite{jiang2024robust}, which regularizes the deformation induced by the learned skinning, and $\gL_{\ell_1}$ promotes sparsity. Despite post-training on only a handful of loose garments data, the model generalizes well across unseen garments and identities (\Cref{fig:deformer-result}).

\vspace{1mm}\noindent
\textbf{Relighting.}
To support relighting of the reconstructed avatars, we use learnable radiance transfer proposed in~\cite{wang2025relightable}. Adding relightability to LCA only requires the replacement of MLP heads for the canonical decoder and pose-dependent decoder. The model is post-trained with time-multiplexed light stage data~\cite{bi2021deep,martinez2024codec}.

\section{Experiments}
\label{section:experiments}
\subsection{Implementation Details}
\noindent\textbf{Preprocessing.} We process each training dataset to compute (i) foreground and face segmentation, (ii) a pixel-aligned body mesh, (iii) facial-expression estimates, and (iv) eye-gaze estimates. We use fine-tuned models for all these tasks based on Sapiens~\cite{khirodkar2024sapiens} for both pretraining and post-training data. We primarily use body pose, expressions and eye-gaze to model pose-dependent behaviors, these signals are highest quality in the post-training data.

\vspace{1mm}\noindent\textbf{Model.}
LCA transformer consists of $L=8$ layers, each token is $D=1024$ dimensional. This results in a model size of $800$M parameters. We process all images at a resolution of $1024\times768$. During training, we randomly sample the number of images $N$ from $[1,4]$. At evaluation, we fix $N=4$. We set the Gaussian-to-token ratio $k=8$ and $G=8192$, resulting in $65{,}536$ Gaussians per avatar.

\vspace{1mm}\noindent\textbf{Training.}
We use the AdamW optimizer~\cite{loshchilov2017decoupled} with a learning rate of $4 \times 10^{-4}$ and weight decay of $0.05$. The learning rate follows a cosine annealing schedule, preceded by a brief linear warm-up. We use mixed-precision training and gradient clipping with a maximum norm of $||\nabla||_2=1.0$. Standard image augmentations like random cropping, scaling, flipping, and photometric distortions are used. For post-training, following~\cite{khirodkar2024sapiens}, we use differential learning rates to maintain generalization, applying lower rates to earlier layers and progressively higher rates to later layers. Specifically, we set the layer-wise decay to $0.65$. The learning rates for both decoders are unchanged.

\vspace{1mm}\noindent\textbf{Evaluation.}
We evaluate on two test sets: capture-studio, and in-the-wild. The capture-studio set contains randomly sampled views from a multi-view setup for $100$ subjects. The in-the-wild set contains monocular videos of $1000$ fully held-out subjects captured under unconstrained conditions. We report L1, LPIPS~\cite{zhang2018unreasonable}, and PSNR~\cite{fardo2016formal}, computed exclusively on human pixels using segmentation masks.

\begin{figure}[b]
    \centering
    \vspace{-0.2in}
    \includegraphics[width=0.95\linewidth]{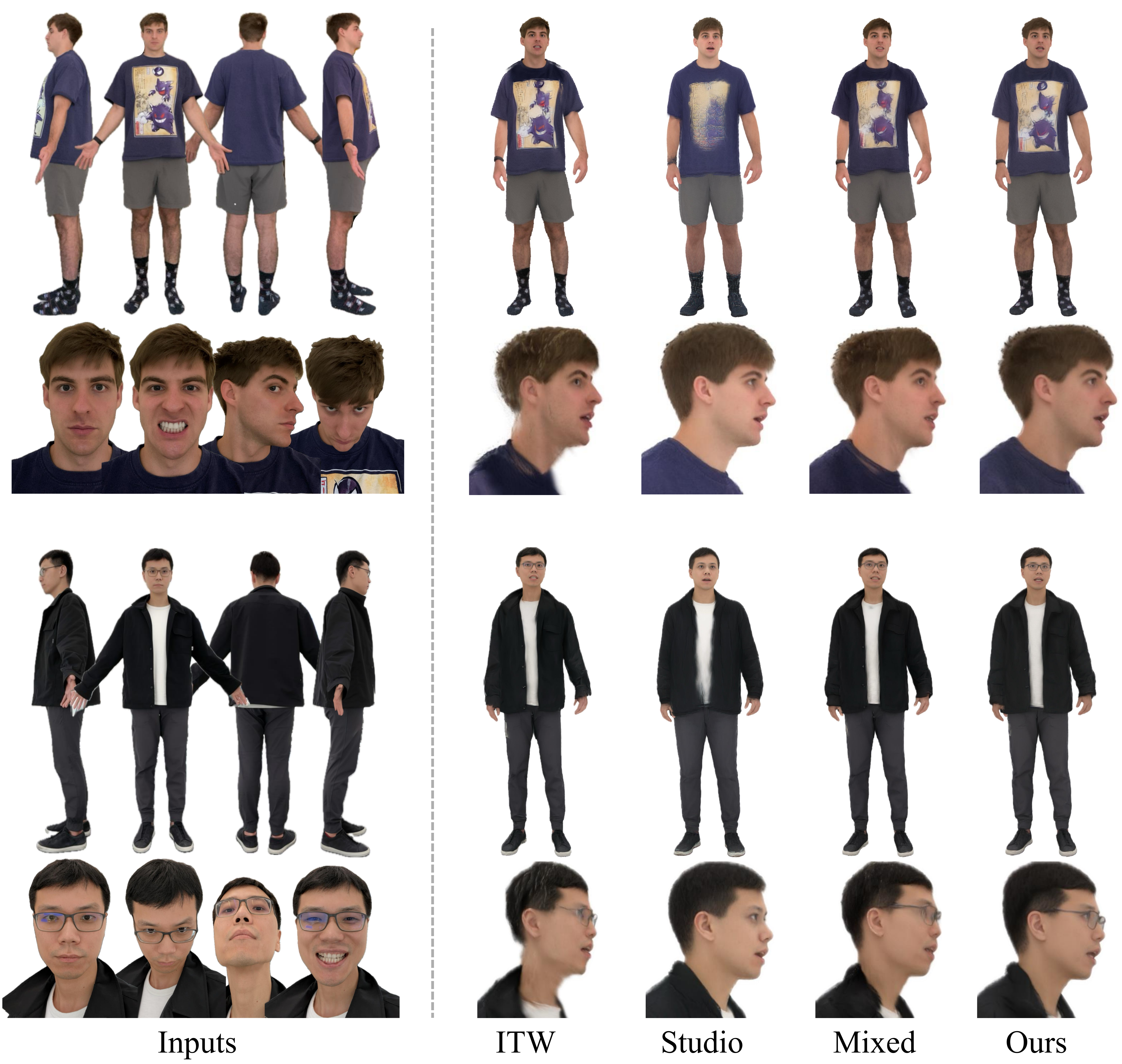}
    \vspace{-0.05in}
    \caption{\textbf{Pretraining vs. Post-Training.} Qualitative comparison of models trained on multiple data sources and training strategies.}
    \vspace{-0.2in}
    \label{fig:pre-post-compare}
\end{figure}

\begin{table}[t]
\centering
\resizebox{\columnwidth}{!}{
\setlength{\tabcolsep}{6pt}
\renewcommand{\arraystretch}{1.3}
\begin{tabular}{c|ccc|ccc}
\toprule
\multirow{2}{*}{\textbf{Train-Source}}
  & \multicolumn{3}{c|}{\textbf{Capture-Studio}}
  & \multicolumn{3}{c}{\textbf{In-the-Wild}} \\
  & L1$\downarrow$      & LPIPS$\downarrow$  & PSNR$\uparrow$
  & L1$\downarrow$      & LPIPS$\downarrow$  & PSNR$\uparrow$ \\
\midrule

In-the-Wild
  & 0.0117    & 0.0904    & 27.464
  & \cellcolor{2nd}{0.0099}    & \cellcolor{2nd}{0.0507}    & 27.669 \\

Capture-Studio
  & 0.0091    & 0.0766    & 29.741
  & 0.0128    & 0.0706    & 26.632 \\

Mixed
  & \cellcolor{2nd}{0.0087}    & \cellcolor{2nd}{0.0724}    & \cellcolor{2nd}{30.025}
  & \cellcolor{2nd}{0.0099}    & 0.0519                      & \cellcolor{2nd}{27.998} \\

\textbf{Pre$\rightarrow$Post (Ours)}
  & \cellcolor{1st}{0.0082}    & \cellcolor{1st}{0.0688}    & \cellcolor{1st}{30.514}
  & \cellcolor{1st}{0.0096}    & \cellcolor{1st}{0.0491}    & \cellcolor{1st}{28.175} \\

\bottomrule
\end{tabular}
}
\vspace{-0.05in}
\caption{Effect of training schemes evaluated across domains.}
\vspace{-0.2in}
\label{tab:preposttrain}
\end{table}

\begin{table}[t]
\centering
\resizebox{\columnwidth}{!}{%
\renewcommand{\arraystretch}{1.2}
\begin{tabular}{c|ccc|ccc}
\toprule
\multirow{2}{*}{\textbf{Method}}
  & \multicolumn{3}{c|}{\textbf{Capture-Studio}}
  & \multicolumn{3}{c}{\textbf{In-the-Wild}}\\
  & L1$\downarrow$      & LPIPS$\downarrow$  & PSNR$\uparrow$
  & L1$\downarrow$      & LPIPS$\downarrow$  & PSNR$\uparrow$ \\
\midrule
\multicolumn{7}{c}{\textbf{(a) Multi-view}}\\
\midrule
ExAvatar~\cite{moon2024expressive}
  & 0.011 & 0.065 & 23.925
  & 0.033 & 0.120 & 18.010 \\

UP2You~\cite{cai2025up2you}
  & 0.021 & 0.085 & 19.585
  & 0.025 & 0.086 & 19.062 \\

MV-LHM*~\cite{LHM}
  & \cellcolor{2nd}{0.008} & \cellcolor{2nd}{0.043} & \cellcolor{2nd}{26.606}
  & \cellcolor{2nd}{0.009} & \cellcolor{2nd}{0.040} & \cellcolor{2nd}{25.796} \\

\textbf{LCA (Ours)}
  & \cellcolor{1st}{0.007} & \cellcolor{1st}{0.041} & \cellcolor{1st}{27.483}
  & \cellcolor{1st}{0.008} & \cellcolor{1st}{0.029} & \cellcolor{1st}{27.802} \\
\midrule
\multicolumn{7}{c}{\textbf{(b) Single-view}}\\
\midrule
LHM~\cite{LHM}
  & 0.016 & 0.072 & 21.897
  & 0.036 & 0.109 & 18.322\\

\textbf{LCA (Ours)}
  & \cellcolor{1st}{0.008} & \cellcolor{1st}{0.043} & \cellcolor{1st}{26.878}
  & \cellcolor{1st}{0.008} & \cellcolor{1st}{0.030} & \cellcolor{1st}{27.685}\\
\bottomrule
\end{tabular}
}
\vspace{-0.05in}
\caption{Quantitative comparison with state-of-the-art 3D avatar methods. * denotes methods trained by us for multi-view inputs.}
\vspace{-0.2in}
\label{tab:sota}
\end{table}

\begin{figure*}[b]
    \centering
    \vspace{-0.1in}
    \includegraphics[width=0.9\linewidth]{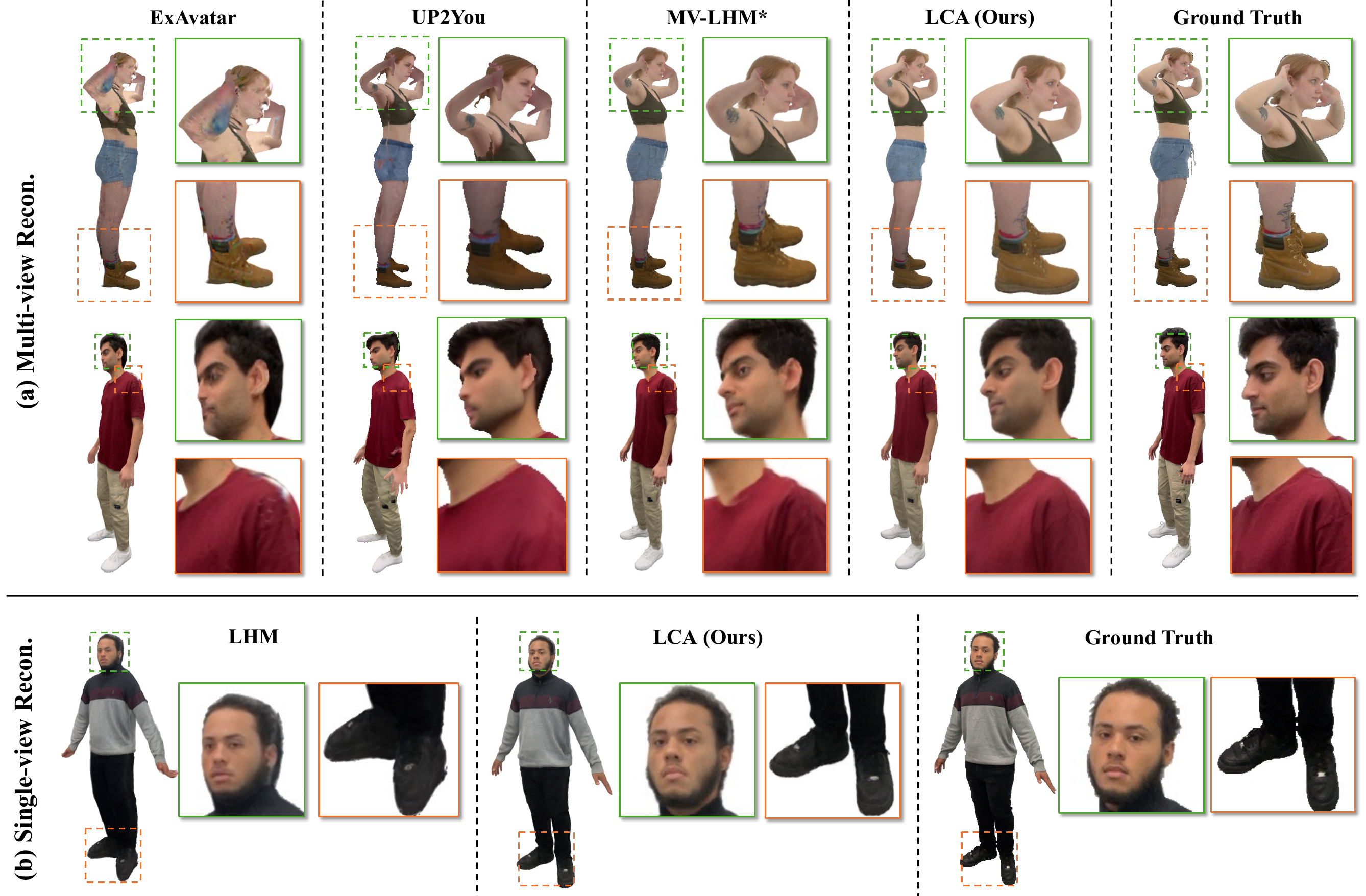}
    \caption{\textbf{Qualitative Comparison with State-of-the-Art Methods.} LCA outperforms in both multi-view and monocular settings.}
    \label{fig:sota}
\end{figure*}

\subsection{Pretraining vs. Post-Training}
We study the effectiveness of large-scale pretraining and post-training across diverse data sources. \Cref{tab:preposttrain} compares models trained only on capture-studio data, only on in-the-wild data, on their mixture, and our proposed pretrain$\rightarrow$post-train scheme. Surprisingly, the studio-only model does not perform best even on the studio domain, suggesting that it overfits to identity appearance and viewpoints. In contrast, a model trained on mixed data provides a strong baseline, achieving about $30.0$ PSNR on studio and $28.0$ PSNR on in-the-wild test set, which mirrors the most common training recipe in existing methods~\cite{LHM, PERSE, sim2025persona}. Our pretrain$\rightarrow$post-train approach, however, outperforms this mixed strategy across domains -- reaching $30.5$ PSNR on studio data highlighting improved realism and $28.2$ PSNR on the in-the-wild set showcasing stronger generalization.

\Cref{fig:pre-post-compare} qualitatively compares feed-forward avatar creation from eight input images for all methods. We show that the in-the-wild-only model produces blurry avatar with incomplete 3D geometry, whereas the studio-only model is sharper and more 3D-complete but fails to generalize to unseen clothing and accessories such as jackets and eyewear. The mixed-data model behaves similarly to the in-the-wild model with slightly better fidelity. Our approach yields sharper, more 3D-complete avatars that generalize better to diverse clothing and accessories.

\subsection{Comparison with State-of-the-Art Methods}
\Cref{tab:sota} compares LCA with existing state-of-the-art methods~\cite{moon2024expressive,cai2025up2you,LHM} using their recommended evaluation protocols, in both multi-view and monocular setups. For fairness, we extend monocular baselines to multi-view setting using our data, \eg, MV-LHM denotes our multi-view extension of LHM~\cite{LHM} trained with the mixture of our pretraining and post-training data. LCA consistently outperforms prior methods across setups and data distributions. In the multi-view studio setting, it surpasses ExAvatar~\cite{moon2024expressive} by $3.56$~dB PSNR, and by $9.8$~dB in the in-the-wild setting. Because ExAvatar~\cite{moon2024expressive} is optimization-based, it does not generalize well to unseen viewpoints during avatar fitting, leading to lower scores overall.
We observe the largest gains on faces and extremities such as fingers and legs. When repurposed for monocular (single-view) input, LCA outperforms LHM~\cite{LHM} by $5.0$~dB PSNR on studio data and by $9.3$~dB PSNR in the in-the-wild setting. Overall, our method produces sharper local details (\eg mouth, fingers) and more faithful articulation than the baselines.

\Cref{fig:sota} provides qualitative comparisons. LCA avatars produce faithful geometry and appearance, whereas other methods often exhibit 3D distortions, especially around the nose and hips. Our avatar also preserves fine appearance details such as shoelaces, benefiting from high-resolution supervision. Despite never being post-trained on examples with eyewear or headwear, LCA generalizes to these, as well as to diverse hairstyles, clothing, ethnicities, and even stylized characters (see~\Cref{figure:teaser}). Overall, LCA produces more expressive facial animations with better preservation of subtle changes such as eye gaze, cheek motion, and inner-mouth details.
We additionally compare with Wan-Animate~\cite{cheng2025wan} (2D video diffusion) and GUAVA~\cite{zhang2025guava} (upper-body 3D avatar) in the supplementary material.

\subsection{Discussion}
\label{sec: Discussion}

\begin{figure}[t]
    \centering
    \includegraphics[width=\linewidth]{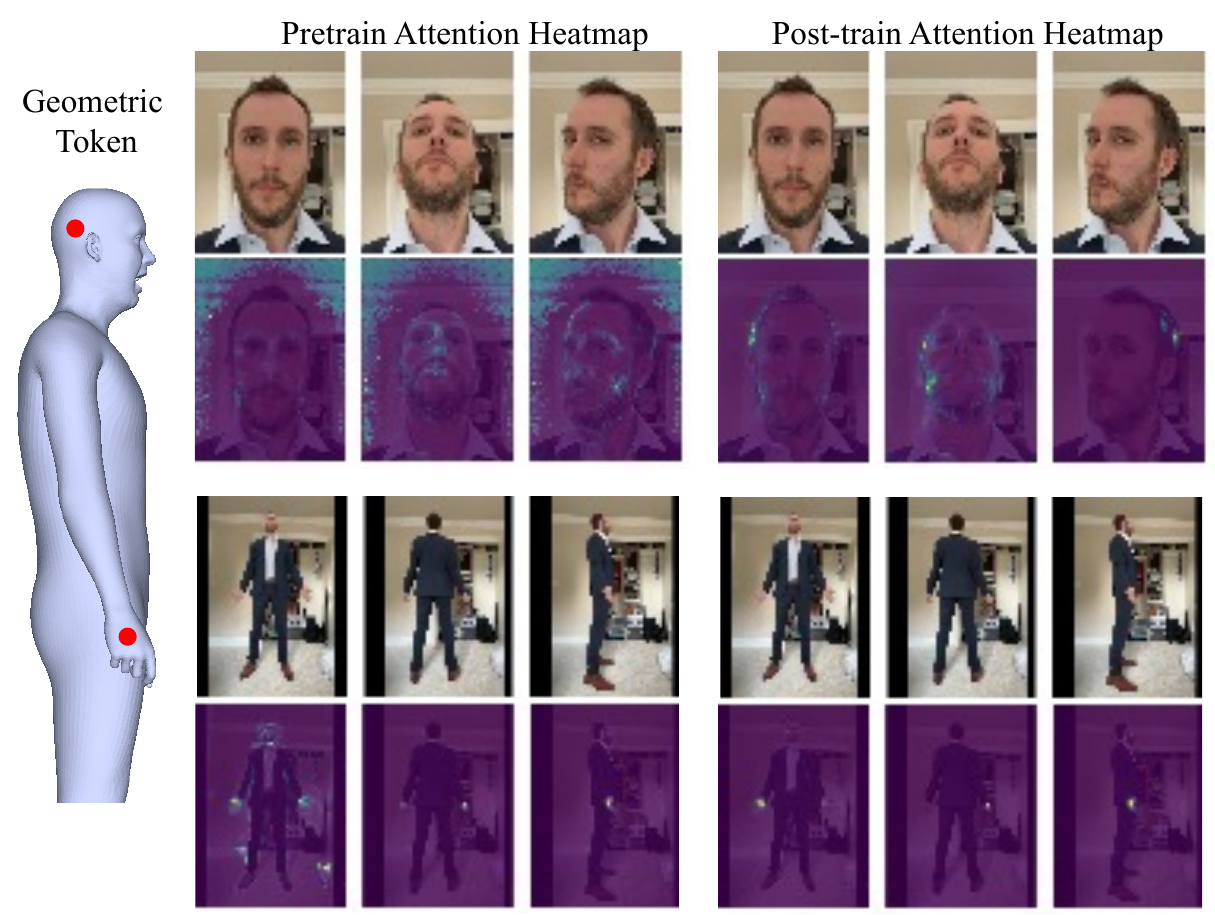}
    \caption{\textbf{Attention Map Visualization.} Post-training yields cleaner semantic correspondences in last-layer attention maps between geometric and image tokens. The selected geometric tokens on the mesh are shown in red.}
    \label{fig:attn-heatmap}
    \vspace{-0.2in}
\end{figure}

\textbf{Effect on Attention Maps.} We show that LCA implicitly learns semantic correspondences between mesh vertices and input images. \Cref{fig:attn-heatmap} visualizes the post-training change in last-layer attention between selected geometric token and image patches, shown as heatmaps. For instance, a vertex on the back of the head attends to image regions corresponding to the head, while a hand vertex attends to pixels around the left hand. Compared to pretraining, post-training reduces the noise in the attention maps, yielding cleaner correspondences with human pixels.

\begin{table}[t]
\centering
\resizebox{\columnwidth}{!}{%
\renewcommand{\arraystretch}{1.3}
\resizebox{\columnwidth}{!}{%
\begin{tabular}{c|ccc|ccc}
\toprule
\multirow{2}{*}{\textbf{\# Videos}} & \multicolumn{3}{c|}{\textbf{Capture-Studio}} & \multicolumn{3}{c}{\textbf{In-the-Wild}} \\
     & L1$\downarrow$      & LPIPS$\downarrow$  & PSNR$\uparrow$
  & L1$\downarrow$      & LPIPS$\downarrow$  & PSNR$\uparrow$    \\ \midrule
$10$K  & 0.0083                 & 0.0681                    & 30.393                    & 0.0109                 & 0.0536                    & 27.665                   \\
$100$K & 0.0083                 & \cellcolor{1st}{0.0677}                    & 30.457                    & 0.0108                 & 0.0533                    & 27.755                   \\
$1$M   & \cellcolor{1st}{0.0082}                 & 0.0688                    & \cellcolor{1st}{30.514}                    & \cellcolor{1st}{0.0096}                 & \cellcolor{1st}{0.0491}                    & \cellcolor{1st}{28.175}                   \\
\bottomrule
\end{tabular}%
}
}
\caption{Effect of scaling pretraining data on downstream performance across data distributions.}
\vspace{-0.2in}
\label{tab:datascale}
\end{table}

\vspace{1mm}\noindent
\textbf{Scaling Pretraining Data.}
We study how the pretraining data scale affects performance. To this end, we subsample the data to $10$K, $100$K, and $1$M videos while keeping all other details fixed. All models are pretrained and post-trained using the same hyperparameters. \Cref{tab:datascale} shows that on the studio benchmark, all three settings achieve similar performance, suggesting that, when combined with multi-view post-training, even a relatively small amount of in-the-wild pretraining suffices for the studio domain. In contrast, performance on the in-the-wild test set depends strongly on pretraining scale: increasing from $10$K to $100$K to $1$M videos significantly reduces reconstruction error.

\begin{figure}[b]
    \centering
    \vspace{-0.1in}
    \includegraphics[width=1.0\linewidth]{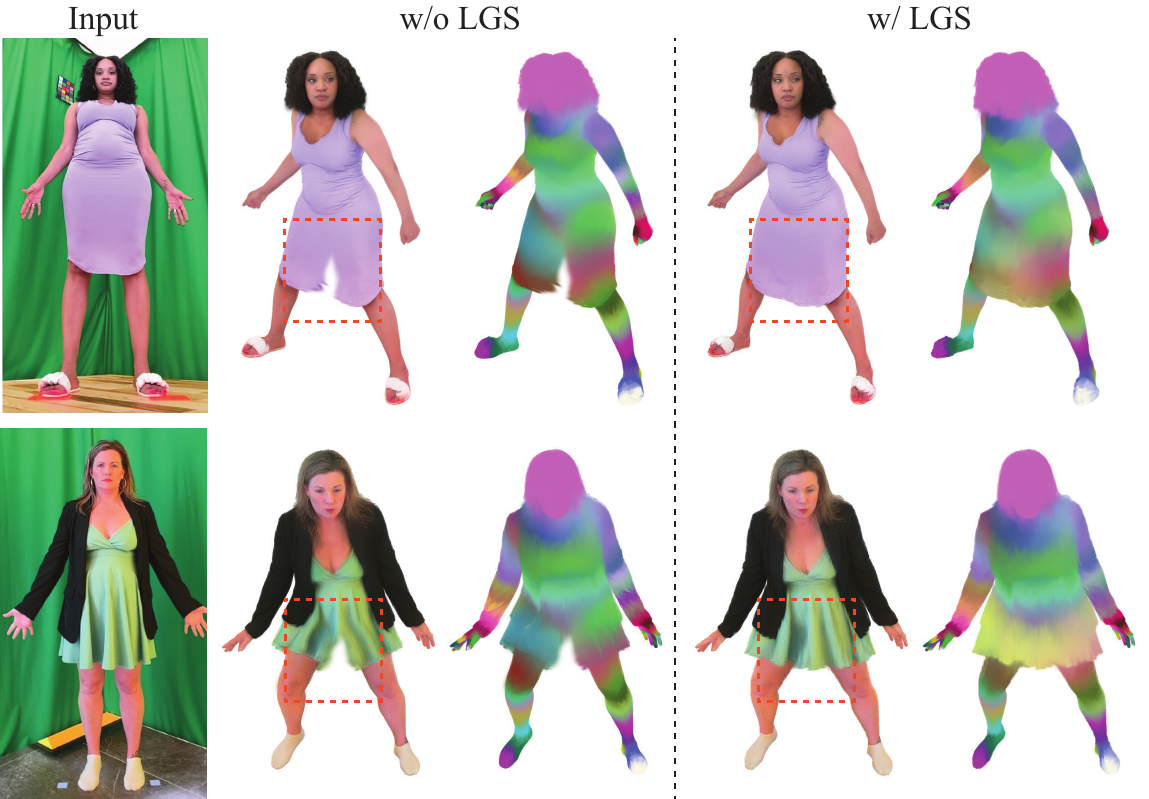}
    \caption{\textbf{Loose Garment Support (LGS).} (\textit{Left}) Frontal view of the input condition. (\textit{Middle}) Post-trained LCA avatar without loose garment support, while the general shape is recovered, skirts behave like pants when moving. (\textit{Right}) LCA with loose garment support produces plausible animations without splitting garments.}
    \label{fig:deformer-result}
\end{figure}

\vspace{1mm}\noindent
\textbf{Loose Garment Support.} We post-train the LCA model with loose garment support on $147$ subjects with loose garments in addition to our existing post-train dataset. As shown in~\Cref{fig:deformer-result}, the original post-trained LCA suffers from splitting leg artifacts due to the predefined skinning weights derived from a minimally clothed body template. In contrast, the modified model more faithfully deforms loose garments without splitting inside the garments even for unseen identities casually captured with a mobile phone.

\begin{figure}[t]
    \centering
    \addtolength{\tabcolsep}{-5.4pt}
    \renewcommand{\arraystretch}{0.3}
    \begin{tabular}{@{}ccccc@{}}
      \includegraphics[width=0.1963\linewidth]{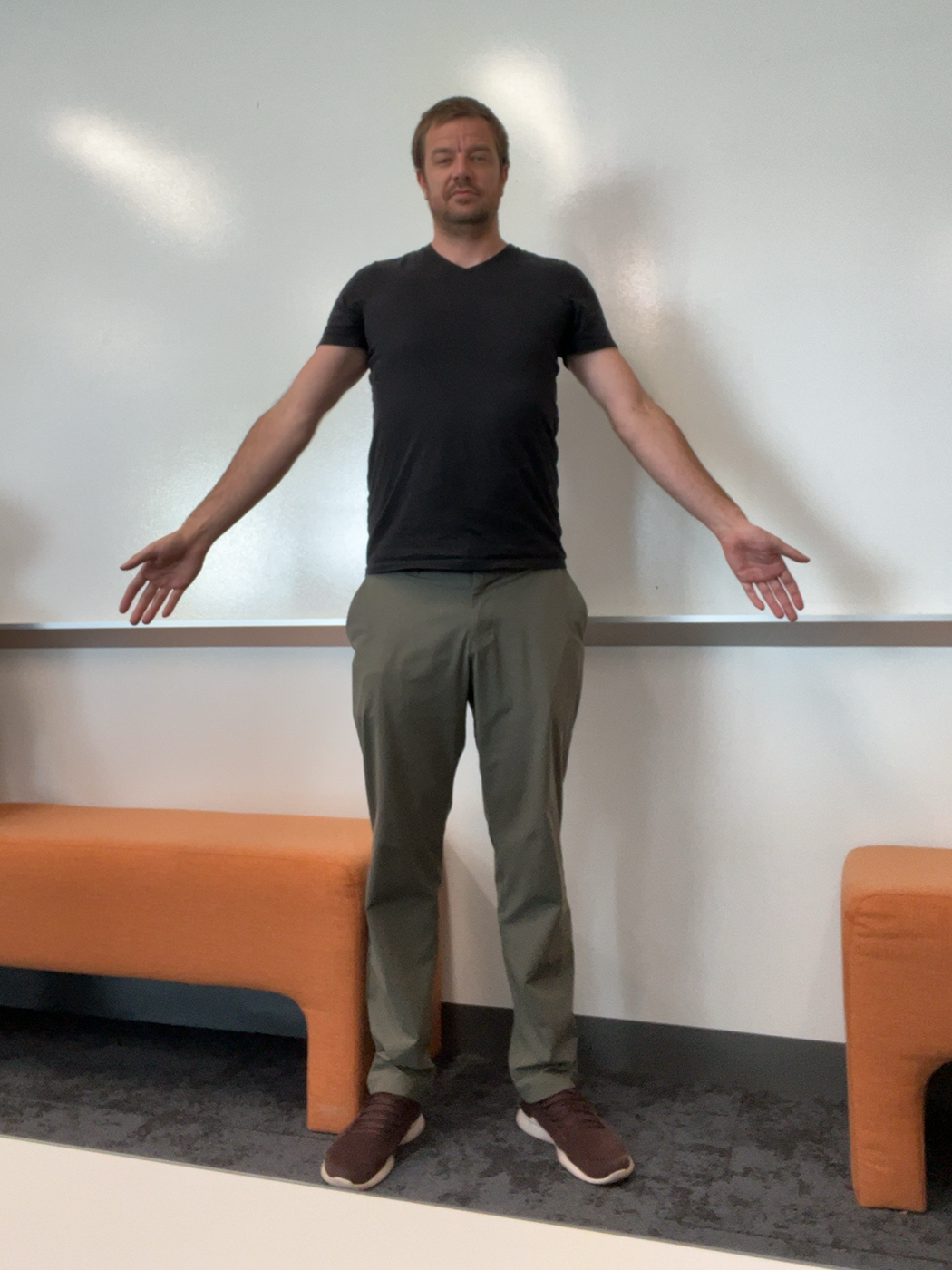} & 
      \includegraphics[width=0.1963\linewidth]{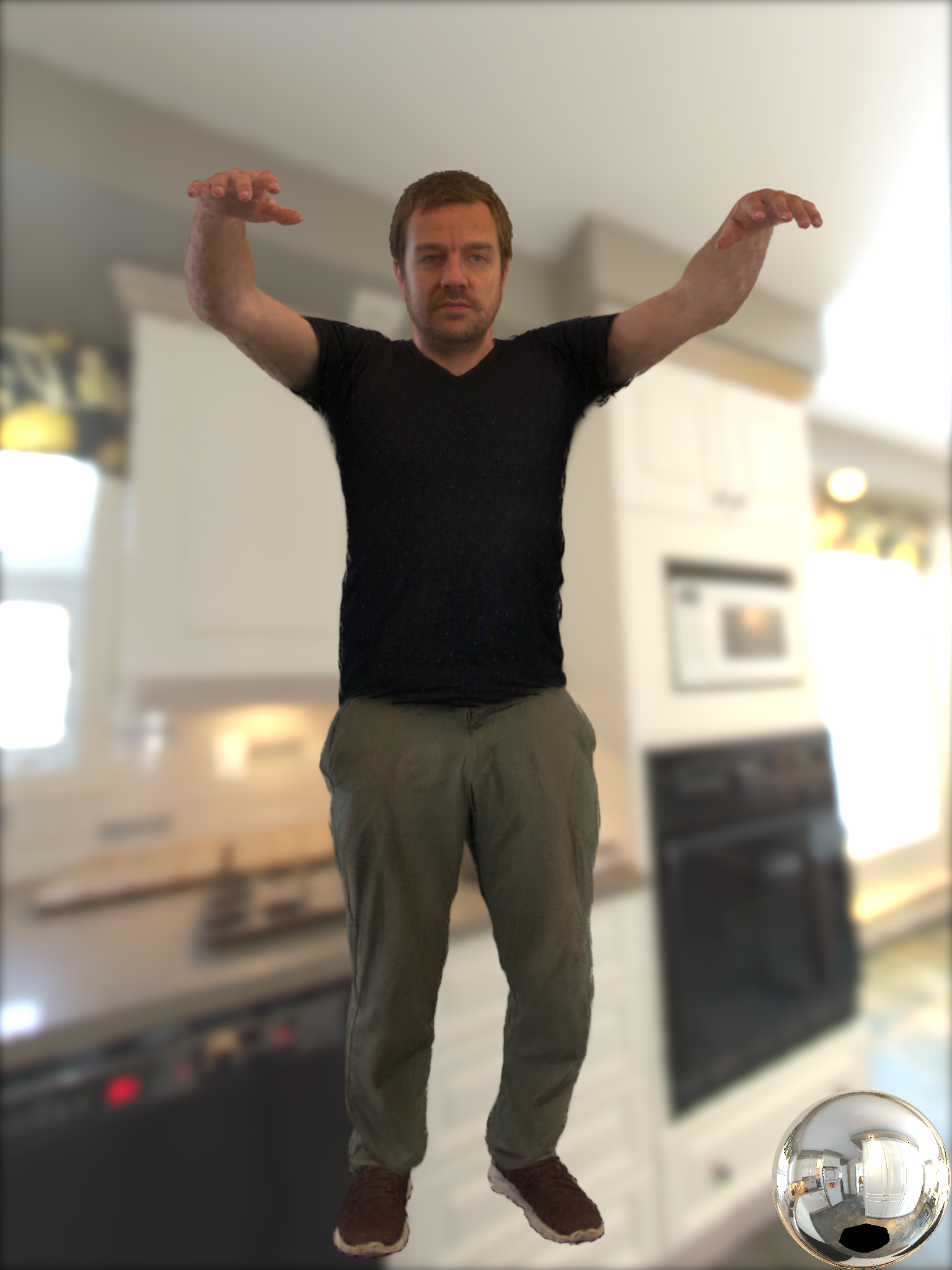} & 
      \includegraphics[width=0.1963\linewidth]{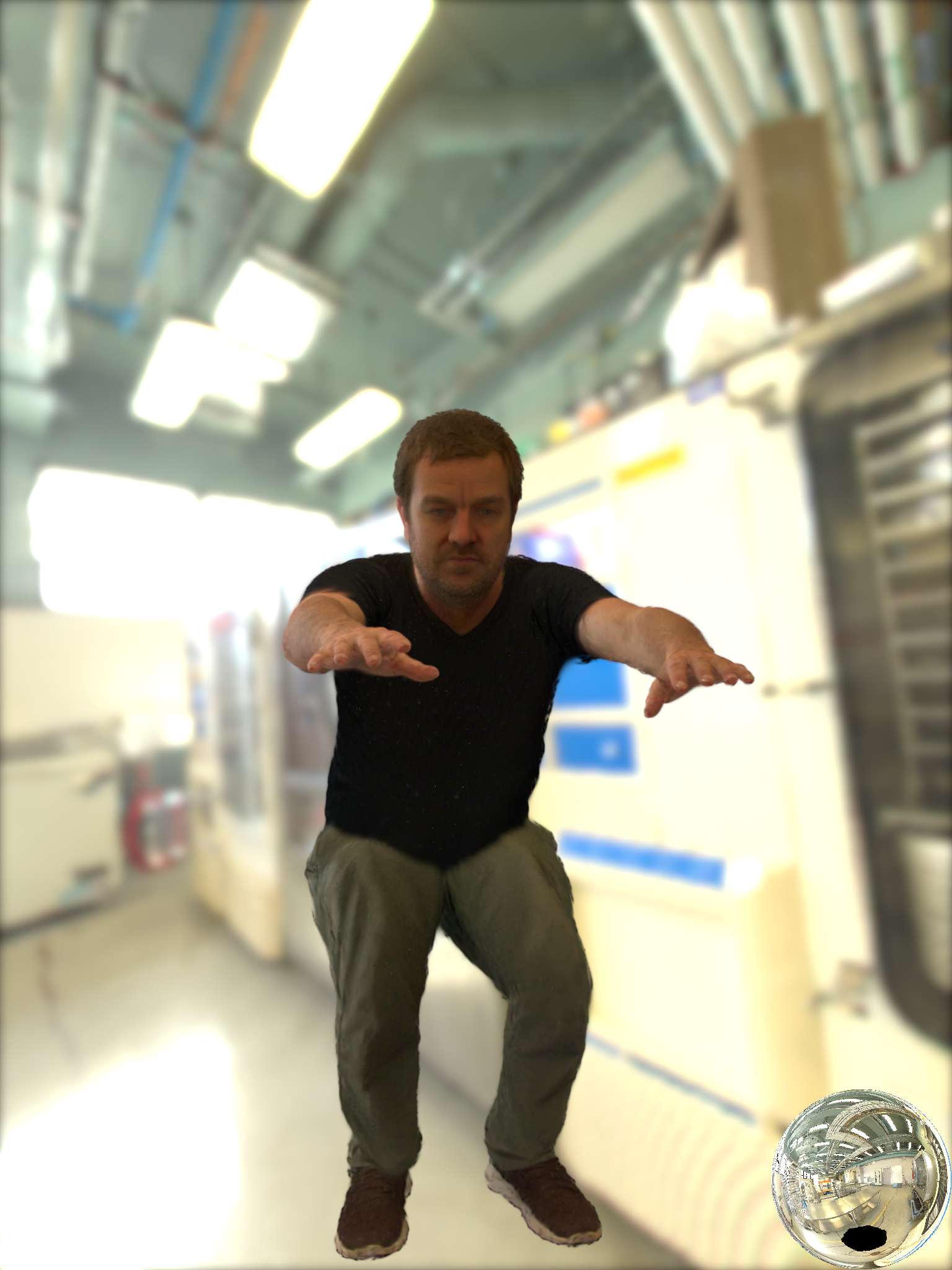} & 
      \includegraphics[width=0.1963\linewidth]{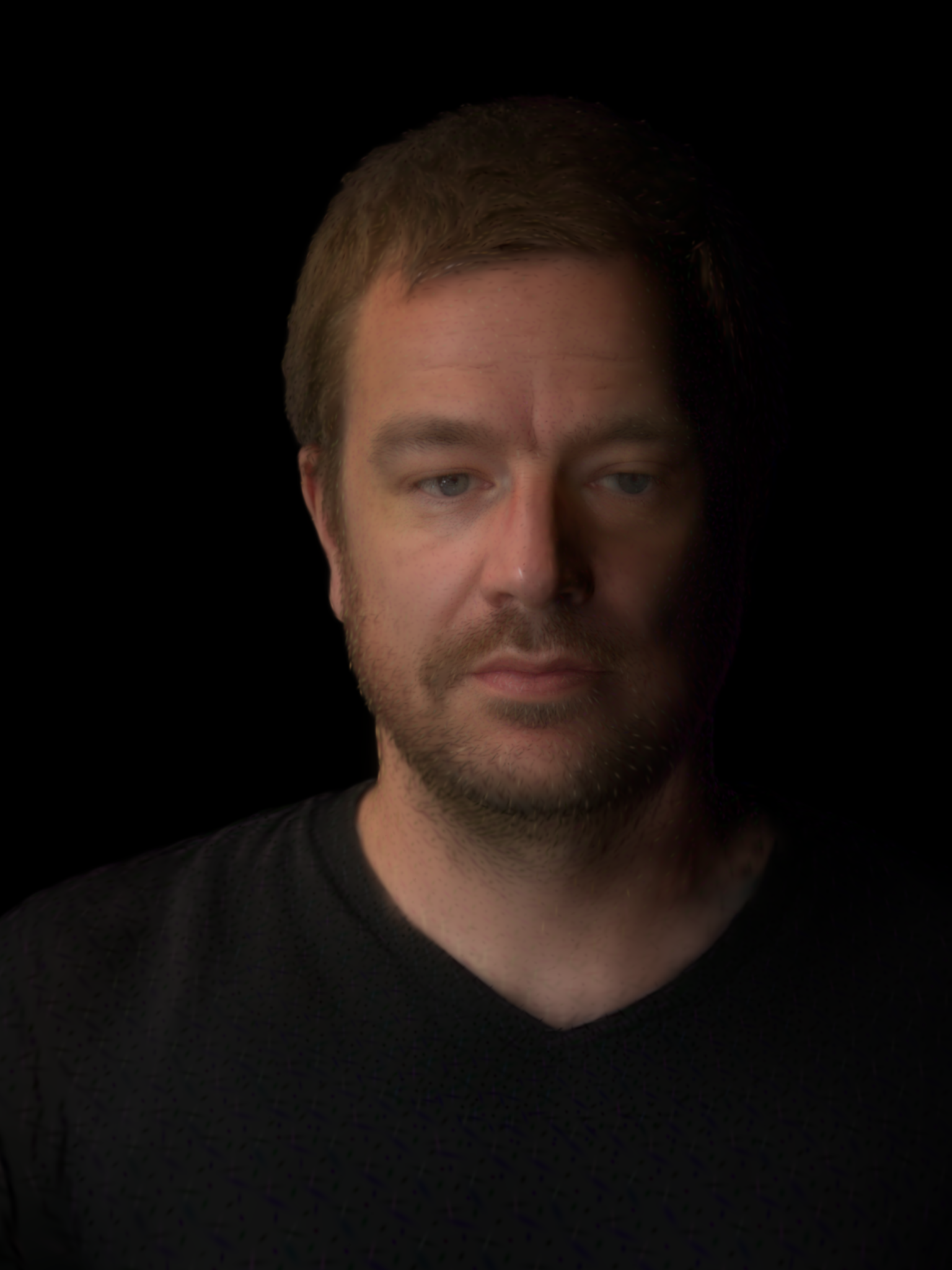} & 
      \includegraphics[width=0.1963\linewidth]{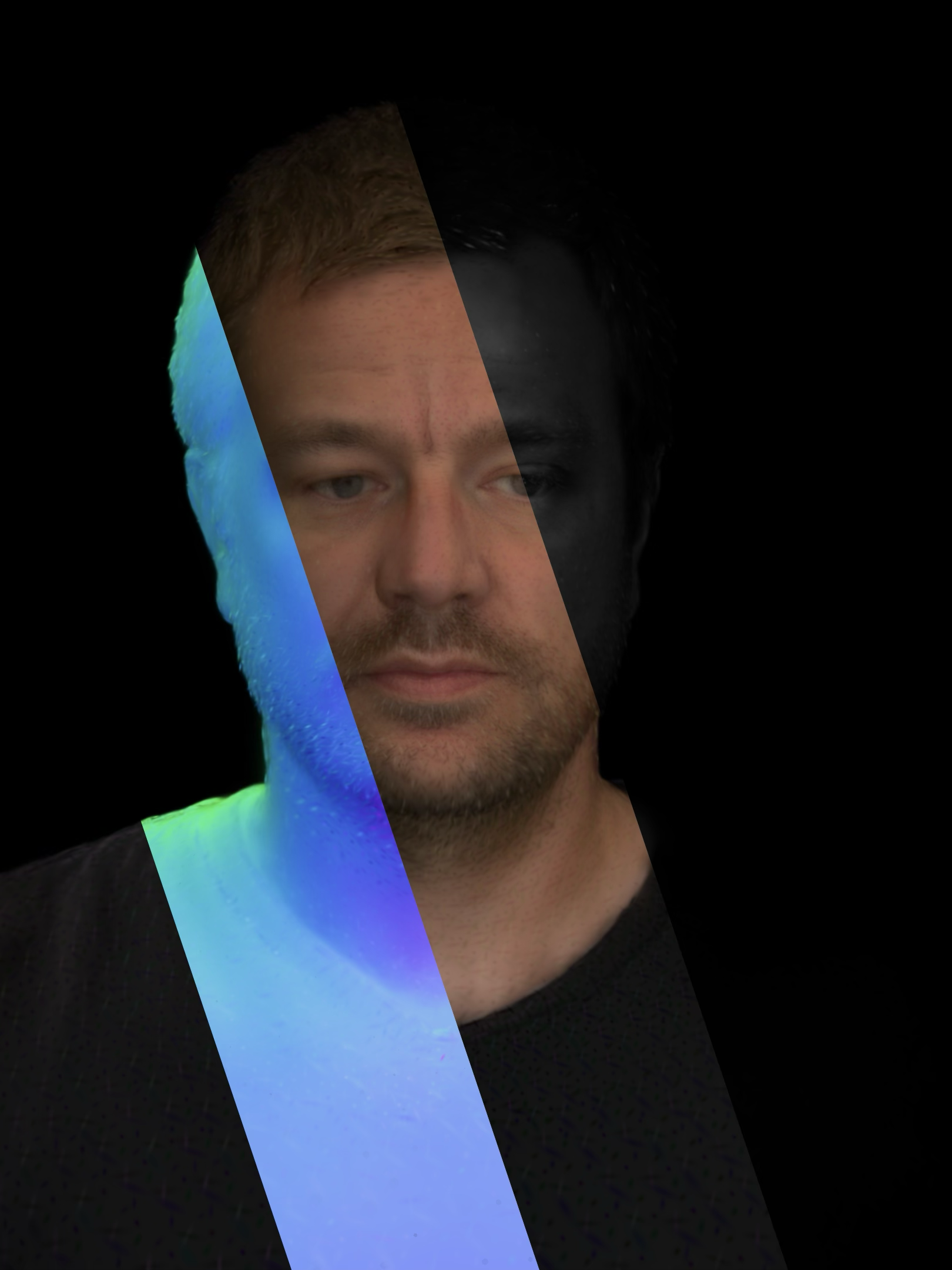} \\
      \includegraphics[width=0.1963\linewidth]{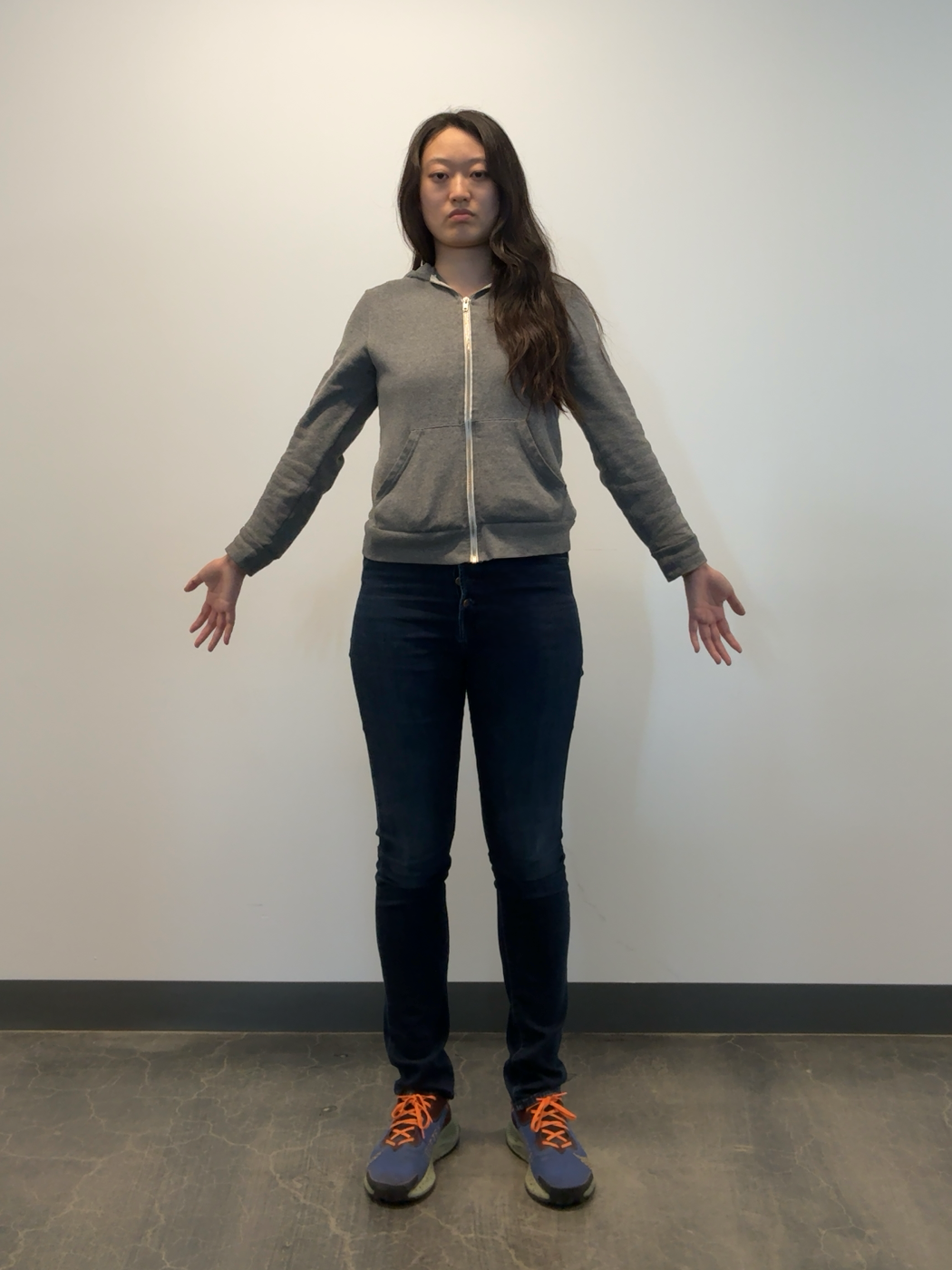} & 
      \includegraphics[width=0.1963\linewidth]{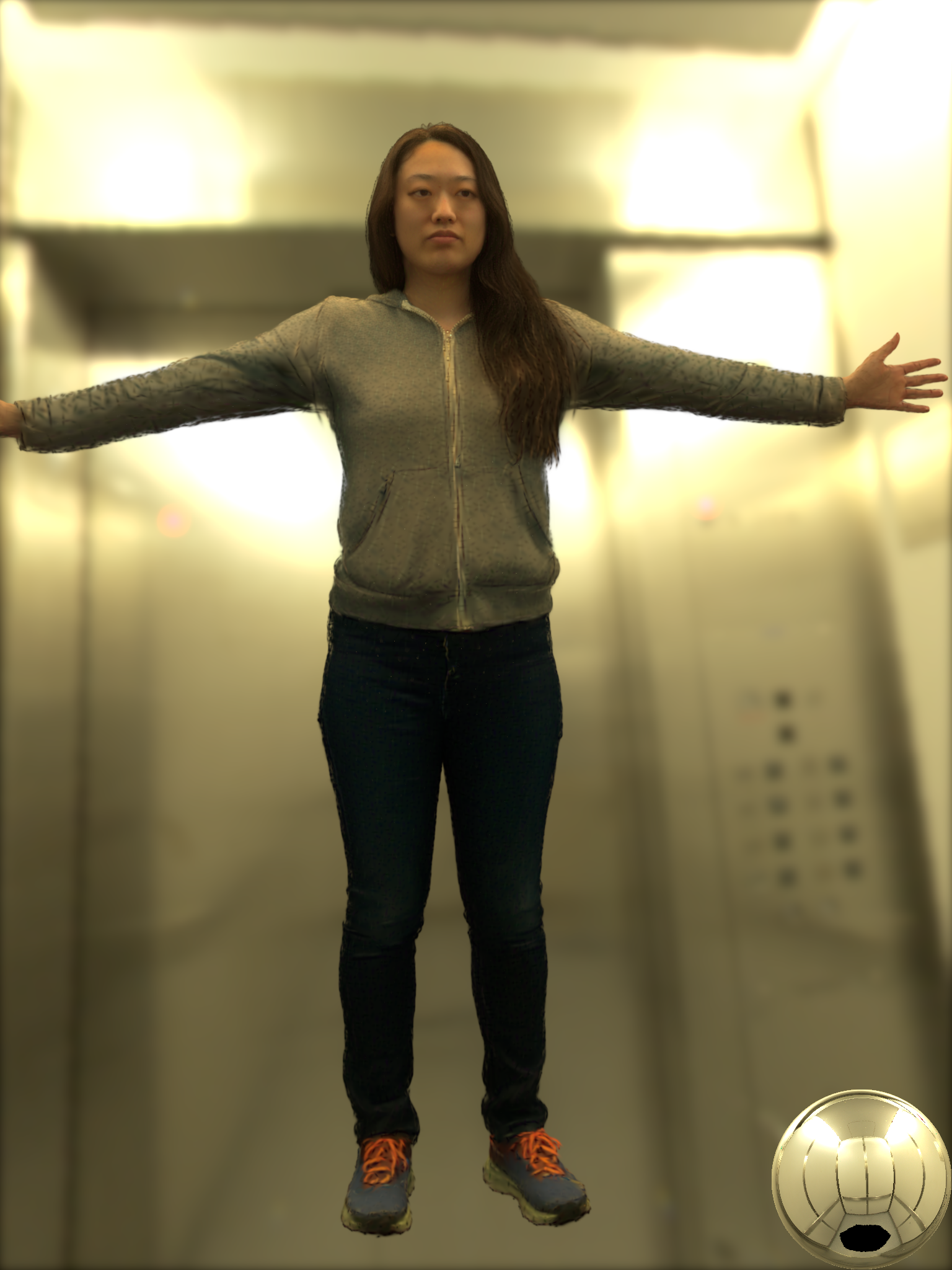} & 
      \includegraphics[width=0.1963\linewidth]{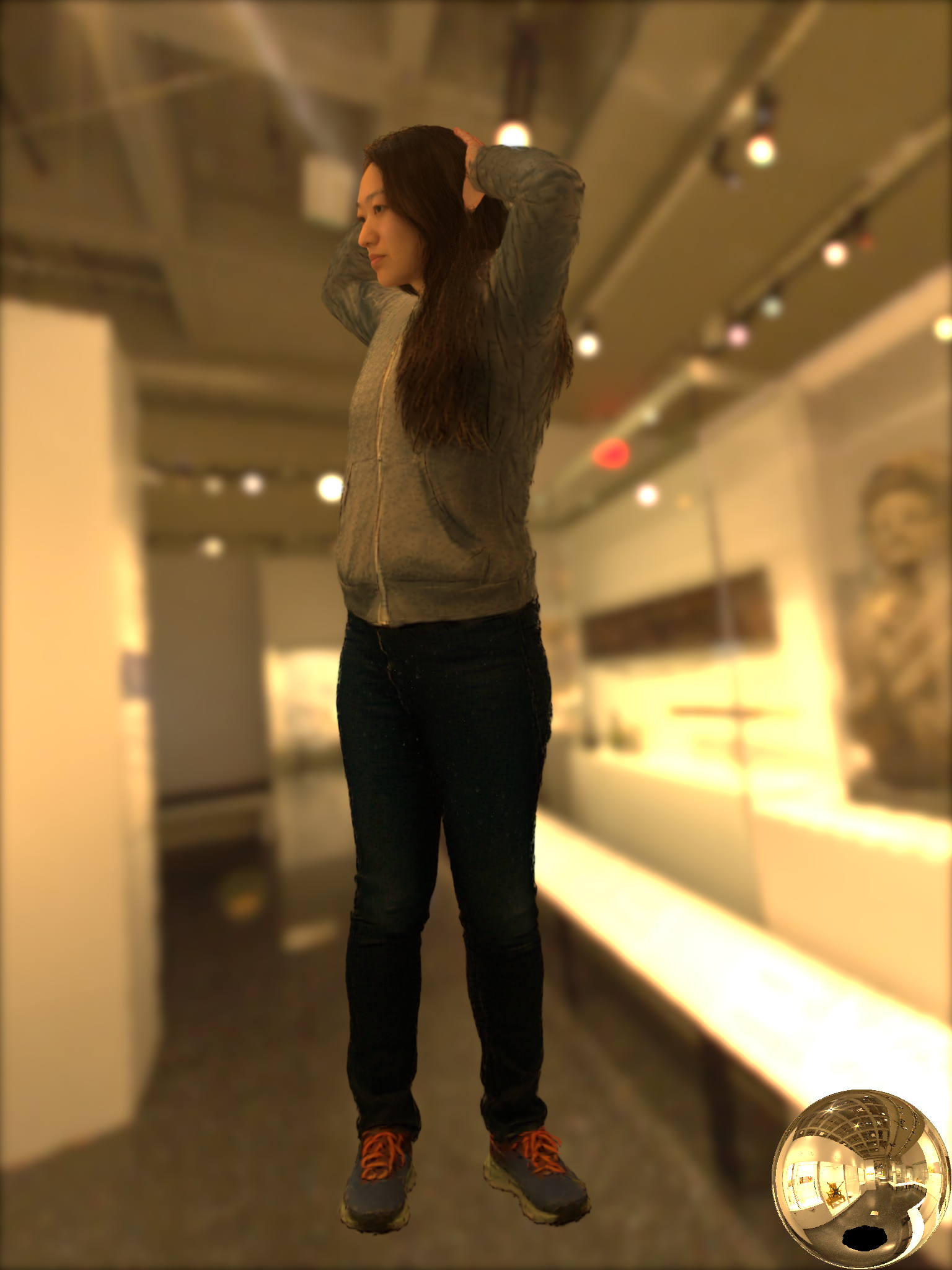} & 
      \includegraphics[width=0.1963\linewidth]{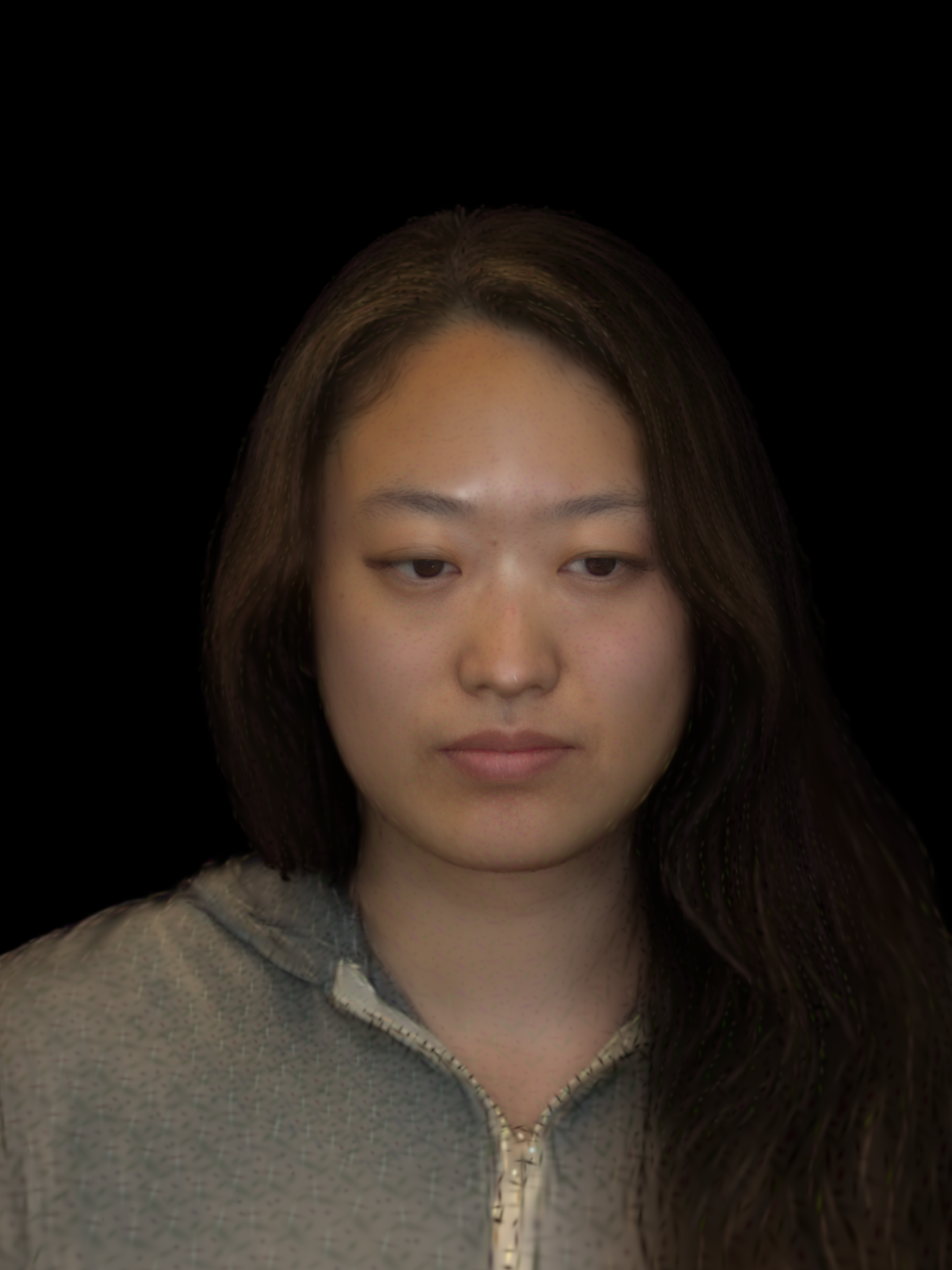} & 
      \includegraphics[width=0.1963\linewidth]{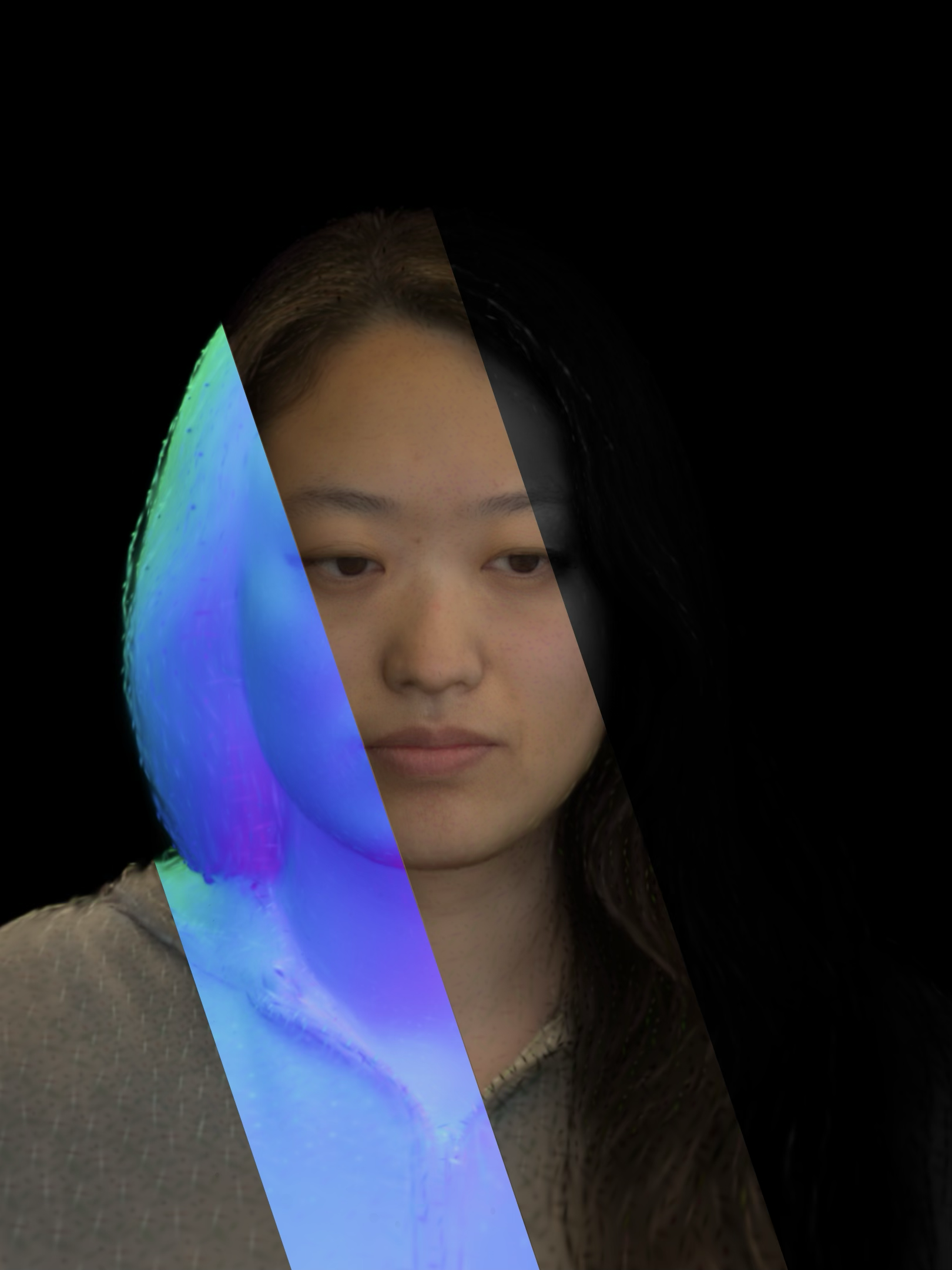} \\ [2pt]
      {\fontsize{8.5pt}{10pt}\selectfont Input} & {\fontsize{8.5pt}{10pt}\selectfont Env. Map $1$} & {\fontsize{8.5pt}{10pt}\selectfont Env. Map $2$} & {\fontsize{8.5pt}{10pt}\selectfont Point Light} & {\fontsize{8.5pt}{10pt}\selectfont Intrinsics} \\ [2pt]
      & & & & {\fontsize{3.5pt}{4.2pt}\selectfont (albedo/normal/diffuse/specular)} \\
    \end{tabular}
    \vspace{-8pt}
    \caption{\textbf{Relightable LCA.} We demonstrate relighting under HDRI environment maps and point lights, alongside the recovered intrinsic properties of the avatar.}
    \label{fig:relighting}
    \vspace{-0.2in}
\end{figure}

\vspace{1mm}\noindent
\textbf{Relighting.}
We post-train our relightable LCA extension on multi-view captures with time-multiplex light patterns including fully-lit and partially-lit frames. As shown in~\Cref{fig:relighting}, LCA recovers plausible reflectance and produces consistent, photorealistic relighting across diverse lighting environments, despite being conditioned only on unconstrained phone captures at test time. We also visualize albedo and normal recovered by LCA.

\vspace{1mm}\noindent
\textbf{Real-Time Drivability.} Our residual design with separate canonical and pose-dependent decoders enables efficient inference. We process input images once with the transformer and canonical decoder, and driving thereafter uses only the lightweight pose-dependent decoder. This yields real-time performance at $586$ FPS on an A100 GPU.

\vspace{1mm}\noindent
\textbf{Limitations.} Finer details, such as clothing with embroidery and intricate textures, remain challenging. Moreover, future work should address secondary motion dynamics such as hair motion and movement of accessories like handbags. The model does not handle heavy occlusions or fast motion blur, which can degrade reconstruction quality.

\section{Conclusion}
\label{sec: Conclusion}
We introduce LCA, a pre/post-training framework for modeling full-body avatars with high-fidelity animation and faithful 3D details. Our experiments show that it is possible to achieve generalization and high-fidelity generation by effectively leveraging large-scale in-the-wild data and high-quality studio data with the proposed two-stage training strategy, similar to frontier language and vision models. Our design is versatile, enabling feature extensions non-trivial to incorporate with large-scale data such as relighting and loose garment handling, and the resulting avatars run in real time. The multi-stage training paves the way for truly scalable high-fidelity avatar creation for everyone.

\section*{Acknowledgments}
We thank Guy Adam, Amol Agrawal, Hernan Badino, Chun-Wei Chan, Yueh-Tung Chen, Shen-Chi Chen, Yuhua Chen, Carol Cheng, Tingfang Du, Itai Druker, Marco Dal Farra, Ryan Frazier, Sidi Fu, Emanuel Garbin, Ke Gao, Liuhao Ge, Eran Guendelman, Chen Guo, Aaqib Habib, Ish Habib, Andrew Hou, Yuta Inoue, Ethan James, Austin James, Fei Jiang, Sam Johnson, Justin Joseph, Anjani Josyula, Song Ju, Kevin Kane, Kai Kang, Thomas Keady, Taylor Koska, Sanjeev Kumar, Jess Kuts, Jianchao Li, Steven Longay, Alex Ma, Kevyn McPhail, Sergiu Munteanu, Conor O'Hollaren, Eli Peker, Sam Pepose, Albert Parra Pozo, Wei Pu, David Rogers, Javier Romero, Igor Santesteban, Michael Schwarz, Yigal Shenkman, Jake Simmons, Tomas Simon, Nir Sopher, Sam Sussman, Autumn Trimble, Harshita Tupili, Julien Valentin, Carlos Vallespi-Gonzalez, Moran Vatelmacher, Kiran Vekaria, Kishore Venkateshan, Simon Venshtain, Harsh Vora, Yimu Wang, Yuzhi Wang, Michael Wu, Longhua Wu, Chengxiang Yin, Jiu Xu, Bo Yang, Shoou-I Yu, and Junchen Zhang for data processing and discussion.

{
    \small
    \bibliographystyle{ieeenat_fullname}
    \bibliography{references}
}
\clearpage
\setcounter{page}{1}
\setcounter{section}{0}        
\maketitlesupplementary
\renewcommand{\thesection}{\Alph{section}}
\renewcommand{\thetable}{S\arabic{table}}
\setcounter{table}{0}
\renewcommand{\thefigure}{S\arabic{figure}}
\setcounter{figure}{0}

\section{Additional Qualitative Results}
Please see supplementary video for additional qualitative results.

\section{Network Architecture}
In this section, we describe the design of our network architecture in detail.

\subsection{Tokenization Details}
We use Sapiens-1B~\cite{khirodkar2024sapiens} as our image feature extractor. Face crops are obtained by detecting face keypoints using Sapiens, computing a bounding box from the keypoints, and cropping and resizing to the target resolution. The $G{=}8{,}192$ geometric tokens are sampled from the surface of a template mesh from MHR~\cite{ferguson2025mhr}, with half sampled on the face region and half on the body to provide higher resolution in the face area. The positional encoder $\gF_{\text{PE}}$ uses fixed Fourier features: given a 3D point, we compute $\sin$ and $\cos$ at $6$ logarithmically spaced frequency bands ($2^0, 2^1, \ldots, 2^5$) and concatenate the result with the raw input, yielding a $39$-dimensional encoding per point. This encoding is then projected to the token dimension $D{=}1024$ via a single MLP layer ($\mathcal{F}_{\text{proj-gs}}$) for use in the subsequent attention layers.

\subsection{Transformer}
As described in Section~3.1 of the main paper, each LCA Transformer layer comprises three components: (1) self-attention over image tokens, (2) self-attention over geometry tokens, and (3) multimodal attention. All attention modules use $16$ heads.

For the image self-attention module $\gA_{\text{image}}$, we follow VGGT~\cite{wang2025vggt} by augmenting the image token sequence with four additional learned registry tokens, which are discarded after the layer. We also apply 2D Rotary Positional Encoding (2D-RoPE) to preserve spatial information within each image.

For the multimodal attention module $\gA_{\text{multimodal}}$, we adopt a two-stage design inspired by the Body–Face MM-T block in LHM~\cite{LHM}. Specifically, face image tokens attend to face geometry tokens first. The resulting face geometry features are concatenated with body geometry tokens, and this combined set is then attended by body image tokens, enabling bidirectional cross-modal interaction.  Formally,
\begin{align}
\rmT^{\text{gs-face}},\rmT^{\text{gs-body}} &= \text{split}(\rmT^{\text{gs}}),  \\
\rmT^{\text{global}} &= \gF_\text{proj} (\text{AvgPool} (\rmT^{\text{body}})), \\
\rmT^{\text{gs-face}}, \rmT^{\text{face}} &= \gA_{\text{MM-T}} (\rmT^{\text{gs-face}}, \rmT^{\text{face}}; \rmT^{\text{global}}) , \\
\rmT^{\text{gs}} &= \text{concat}(\rmT^{\text{gs-face}},\rmT^{\text{gs-body}}),  \\
\rmT^{\text{gs}}, \rmT^{\text{body}} &= \gA_{\text{MM-T}} (\rmT^{\text{gs}}, \rmT^{\text{body}}; \rmT^{\text{global}}) .
\end{align}
We compute the global feature $\rmT^{\text{global}} \in \R^{1 \times D}$ by first averaging the body image tokens across all spatial locations, followed by a learnable projection through an MLP $\gF_\text{proj}$.

\subsection{Gaussian Decoder}
Both the canonical decoder $\gH_{\text{cano}}$ and the pose-dependent decoder $\gH_{\text{pose}}$ are lightweight networks, each composed of four fully connected layers. We use LeakyReLU activation functions between layers to improve stability and gradient flow. The hidden dimensionality of all intermediate layers is set to 128. This compact design enables fast inference while maintaining sufficient representational capacity for high-quality Gaussian decoding.

\subsection{Inference Efficiency}
To achieve real-time performance, we decouple the inference process into a one-time initialization stage and a runtime animation stage. The computationally intensive transformer encoder and canonical decoder are executed once per subject to generate the canonical Gaussian parameters and geometry tokens. This initialization step takes approximately $2.1$ seconds on a single NVIDIA A100 GPU.

For subsequent animation frames, only the lightweight pose-dependent decoder, $\mathcal{H}_{\text{pose}}$, is evaluated. This component is highly efficient, requiring approximately $1.7$ ms per forward pass. This design ensures that our method allows for high-fidelity, interactive applications such as VR/AR telepresence and real-time character control.

\section{Training Parameters}
We train our models using 64 NVIDIA A100 GPUs (80GB) via Distributed Data Parallel (DDP). For both training stages, we set the per-GPU batch size to 1, resulting in a total effective batch size of 64. The pretraining stage is conducted for $1 \times 10^{5}$ iterations, taking approximately 100 hours. Subsequently, the post-training stage is fine-tuned for $1 \times 10^{4}$ iterations, which requires approximately 10 hours to converge. For the loss weights, we set the $\ell_1$ and LPIPS coefficients to $0.1$ each, and the regularization weight $\lambda = 1.0$.

\section{Analysis of Latent Feature Distribution}
\begin{figure}[h]
    \centering
    \includegraphics[width=\linewidth]{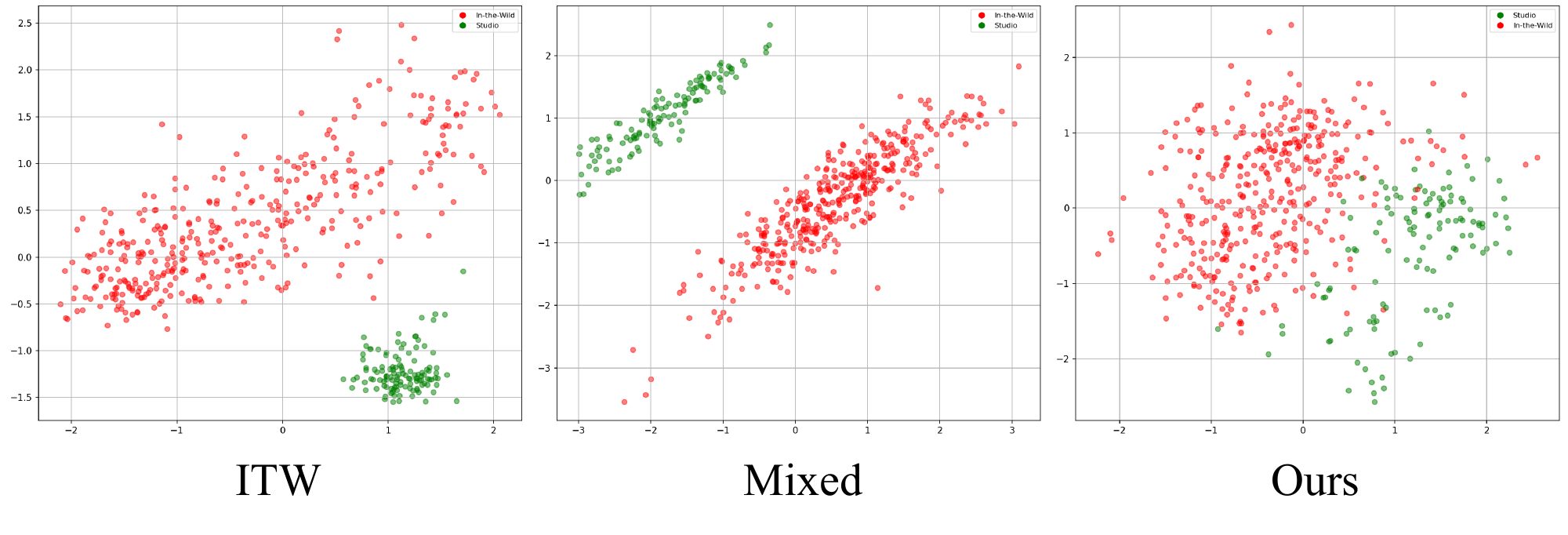}
    \caption{\textbf{PCA of Geometric Token Features.} Visualization of the feature space distributions produced by models trained with different strategies. Green points denote studio-captured subjects, while red points denote in-the-wild subjects.}
    \label{fig:fig-supp-gstoken-pca}
\end{figure}

We analyze the distribution of the learned geometric token features, $T^{gs}$, to understand how different training strategies handle the domain gap between datasets. We extract features for unseen subjects from both the studio-capture and in-the-wild test sets. For each subject, we compute a global feature vector by averaging the geometric tokens and projecting them into 2D space using Principal Component Analysis (PCA). As shown in Figure~\ref{fig:fig-supp-gstoken-pca}, models trained on ITW-only or Mixed data form distinct clusters for studio (green) and in-the-wild (red) samples, limiting synergetic improvement of both fidelity and generalization.  In contrast, our proposed pre/post-training strategy effectively aligns these distributions, treating inputs from both domains consistently. This suggests that our approach learns a robust, high-fidelity, and domain-agnostic representation of human geometry, effectively leveraging the two distinctive training data sources.

\section{Additional Ablation Studies}

\begin{table}[h]
\centering
\resizebox{\columnwidth}{!}{
\setlength{\tabcolsep}{6pt}
\renewcommand{\arraystretch}{1.3}
\begin{tabular}{l|ccc|ccc}
\toprule
\multirow{2}{*}{\textbf{Configuration}}
& \multicolumn{3}{c|}{\textbf{Capture-Studio}}
& \multicolumn{3}{c}{\textbf{In-the-Wild}} \\
& L1$\downarrow$      & LPIPS$\downarrow$  & PSNR$\uparrow$
& L1$\downarrow$      & LPIPS$\downarrow$  & PSNR$\uparrow$ \\
\midrule

\multicolumn{7}{c}{Decoder Architecture} \\
\midrule

Single-Branch
& 0.0088    & 0.0742    & 30.370
& 0.0103    & 0.0516    & 27.936 \\

Dual-Branch (Ours)
& \textbf{0.0082}    & \textbf{0.0688}    & \textbf{30.514}
& \textbf{0.0096}    & \textbf{0.0491}    & \textbf{28.175} \\

\midrule
\multicolumn{7}{c}{Post-Training Learning Rate Decay ($\gamma$)} \\
\midrule
$\gamma = 0.00$
& 0.0117    & 0.0904    & 27.464
& \cellcolor{1st}{0.0096}    & 0.0507    & 27.669 \\

$\gamma = 0.30$
& \cellcolor{1st}{0.0076}    & \cellcolor{1st}{0.0614}    & \cellcolor{1st}{30.609}
& \cellcolor{1st}{0.0096}    & \cellcolor{1st}{0.0482}    & \cellcolor{1st}{28.252} \\

$\gamma = 0.65$ (Ours)
& \cellcolor{2nd}{0.0082}    & \cellcolor{2nd}{0.0688}    & \cellcolor{2nd}{30.514}
& \cellcolor{1st}{0.0096}    & \cellcolor{2nd}{0.0491}    & \cellcolor{2nd}{28.175} \\

$\gamma = 1.00$
& 0.0085    & 0.0694    & 30.464
& 0.0114    & 0.0512    & 27.478 \\

\bottomrule
\end{tabular}
}
\caption{\textbf{Ablation study on decoder architecture and post-training learning rate decay.} Our dual-branch residual design outperforms a single-branch variant. The learning rate decay is critical for preserving pretraining knowledge, with $\gamma{=}0.00$ (no decay) severely degrading studio performance.}
\label{tab:supp_ablations}
\end{table}

We ablate the decoder architecture and post-training learning rate decay ($\gamma$) in \cref{tab:supp_ablations}. Our dual-branch residual design outperforms the single-branch non-residual variant, likely due to improved pose-dependent decoupling. For the learning rate decay, $\gamma{=}0.00$ (no decay, all layers trained at the same rate) severely degrades studio metrics, indicating catastrophic forgetting of pretraining knowledge. Both $\gamma{=}0.30$ and $\gamma{=}0.65$ perform well; we use $\gamma{=}0.65$ in our final model as it offers a good balance across both domains.

\begin{table}[h]
\centering
\resizebox{\columnwidth}{!}{
\setlength{\tabcolsep}{6pt}
\renewcommand{\arraystretch}{1.3}
\begin{tabular}{l|ccc|ccc}
\toprule
\multirow{2}{*}{\begin{tabular}[c]{@{}l@{}}\textbf{Train Data Size}\\[-0.2mm]\small Num. Identities\end{tabular}}
& \multicolumn{3}{c|}{\textbf{Capture-Studio}}
& \multicolumn{3}{c}{\textbf{In-the-Wild}} \\
& L1$\downarrow$ & LPIPS$\downarrow$ & PSNR$\uparrow$
& L1$\downarrow$ & LPIPS$\downarrow$ & PSNR$\uparrow$ \\
\midrule
Pre-100K + Post-500
& 0.0088    & 0.0723    & 30.398
& 0.0130    & 0.0606    & 26.754 \\
Pre-1M + Post-2.7K (Ours)
& \textbf{0.0082}    & \textbf{0.0688}    & \textbf{30.514}
& \textbf{0.0096}    & \textbf{0.0491}    & \textbf{28.175} \\
\bottomrule
\end{tabular}
}
\caption{\textbf{Effect of training data scale.} Pre/post-training benefits persist even at $10\times$ smaller scale (100K pretraining identities, 500 post-training identities), with consistent trends across both domains.}
\label{tab:supp_data_size}
\end{table}

We also study the effect of data scale on the pre/post-training paradigm (\cref{tab:supp_data_size}). Training LCA at $10\times$ smaller scale (100K pretraining identities and 500 post-training identities) still yields improvements over baselines, confirming that the benefits of our two-stage approach are not solely attributable to data scale.

\begin{table}[h]
\centering
\resizebox{0.55\columnwidth}{!}{
\setlength{\tabcolsep}{6pt}
\renewcommand{\arraystretch}{1.3}
\begin{tabular}{l|ccc}
\toprule
\textbf{Method}
& L1$\downarrow$ & LPIPS$\downarrow$ & PSNR$\uparrow$ \\
\midrule
No Deformer
& 0.0304    & 0.2452    & 24.538 \\
LCA (Ours)
& \textbf{0.0238}    & \textbf{0.2189}    & \textbf{26.013} \\
\bottomrule
\end{tabular}
}
\caption{\textbf{Loose garment deformer ablation.} Quantitative evaluation on loose-garment sequences. The full model with the deformer improves all metrics and reduces splitting artifacts.}
\label{tab:supp_deformer}
\end{table}

We quantitatively evaluate the effect of the deformer module on loose-garment sequences in \cref{tab:supp_deformer}. The full model with the deformer improves all metrics and significantly reduces splitting artifacts. We find the deformer benefits from multi-view post-training data to effectively constrain cloth deformation.

\section{Comparison with Alternative Paradigms}
\begin{figure}[h]
    \centering
    \includegraphics[width=0.9\linewidth]{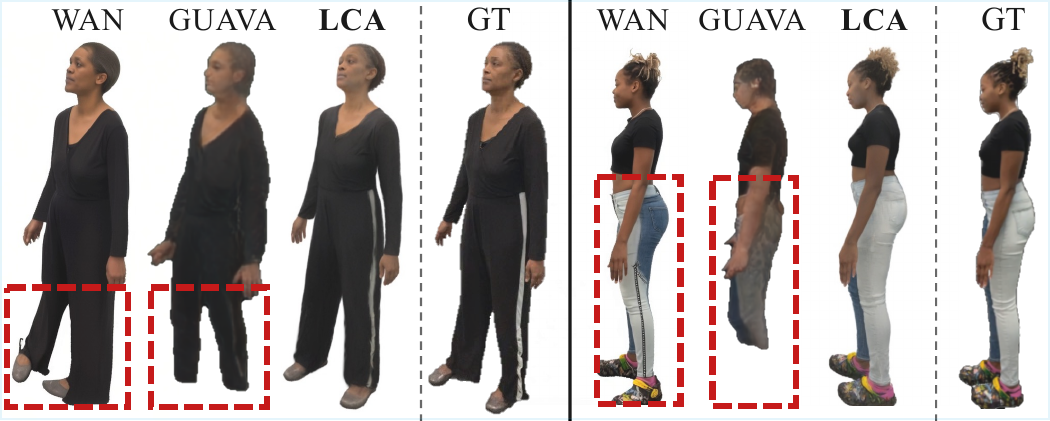}
    \caption{\textbf{Qualitative Comparison with Alternative Paradigms.} Comparison with Wan-Animate~\cite{cheng2025wan} (2D video diffusion) and GUAVA~\cite{zhang2025guava} (upper-body 3D Gaussian avatar).}
    \label{fig:supp-baseline-comparisons}
\end{figure}

We qualitatively compare LCA with Wan-Animate~\cite{cheng2025wan}, a 2D video diffusion method, and GUAVA~\cite{zhang2025guava}, an upper-body 3D Gaussian avatar method, in \cref{fig:supp-baseline-comparisons}. Relative to video diffusion approaches, LCA is more compute-efficient, enables real-time on-device animation, and exhibits stronger holistic 3D consistency, while diffusion baselines can hallucinate and struggle with long-form generation. Compared to GUAVA, which only supports upper-body reconstruction, LCA produces full-body avatars with higher fidelity.

\section{Author Contributions}

\subsubsection*{Project Lead} Shunsuke Saito.

\subsubsection*{Core Contributors} Junxuan Li and Rawal Khirodkar.

\subsubsection*{Loose Garment Support} Zhongshi Jiang, Lingchen Yang, and Rinat Abdrashitov.

\subsubsection*{Relighting} Giljoo Nam and Chengan He.

\subsubsection*{Evaluation \& Benchmarking} Jihyun Lee.

\subsubsection*{Research Discussions} Egor Zakharov, Abhishek Kar, Christian H\"{a}ne, Sofien Bouaziz, Jason Saragih, Yaser Sheikh, and Chen Guo.

\subsubsection*{Data Pipeline \& Processing} {\footnotesize Contributors listed alphabetically by last name.}

\noindent\textit{Pretraining:} Jean-Charles Bazin, James Booth, Wyatt Borsos, Yuan Dong, Peihong Guo, Gin\'{e}s Hidalgo, Matthew Hu, Xiaowen Ma, Julieta Martinez, Marco Pesavento, Yu Rong, Takaaki Shiratori, Carsten Stoll, Zhaoen Su, Anjali Thakrar, Sairanjith Thalanki, Lucy Wang, He Wen, Yichen Xu, and Ariyan Zarei.

\noindent\textit{Post-Training:} Guy Adam, Amol Agrawal, Hernan Badino, Chen Cao, Chun-Wei Chan, Yueh-Tung Chen, Shen-Chi Chen, Yuhua Chen, Carol Cheng, Teng Deng, Tingfang Du, Itai Druker, Marco Dal Farra, Ryan Frazier, Sidi Fu, Emanuel Garbin, Ke Gao, Liuhao Ge, Eran Guendelman, Aaqib Habib, Ish Habib, Xuhua Huang, Yuta Inoue, Ethan James, Sam Johnson, Justin Joseph, Anjani Josyula, Song Ju, Kevin Kane, Kai Kang, Thomas Keady, Taylor Koska, Sanjeev Kumar, Jess Kuts, Jianchao Li, Kai Li, Steven Longay, Kevyn McPhail, Sergiu Munteanu, Eli Peker, Sam Pepose, Albert Parra Pozo, Wei Pu, David Rogers, Javier Romero, Igor Santesteban, Michael Schwarz, Yigal Shenkman, Jake Simmons, Tomas Simon, Nir Sopher, Sam Sussman, Qingyang Tan, Autumn Trimble, Harshita Tupili, Julien Valentin, Carlos Vallespi-Gonzalez, Moran Vatelmacher, Kiran Vekaria, Kishore Venkateshan, Simon Venshtain, Harsh Vora, Yimu Wang, Yuzhi Wang, Michael Wu, Longhua Wu, Jiu Xu, Bo Yang, Chengxiang Yin, Shoou-I Yu, and Junchen Zhang.

\noindent\textit{Evaluation Data:} Andrew Hou, Austin James, Fei Jiang, Alex Ma, and Conor O'Hollaren.

\end{document}